\documentclass[journal,twoside,web]{ieeecolor}
\let\labelindent\relax
\usepackage{tmi}
\usepackage{cite}
\usepackage{amsmath,amssymb,amsfonts}
\usepackage{algorithmic}
\usepackage{graphicx}
\usepackage{textcomp}

\usepackage{latexsym}
\usepackage{multicol} 
\usepackage{changepage}
\usepackage{subfigure}
\usepackage{float}

\usepackage{hyperref}
\usepackage{boldline}
\usepackage{algorithm}
\usepackage{algorithmic}
\usepackage{multirow,array,booktabs}
\usepackage{enumitem}
\usepackage{changepage}
\usepackage{tabularx}
\usepackage[normalem]{ulem}
\usepackage{verbatim}
\usepackage{subcaption}

\usepackage{color, xcolor}
\definecolor{DarkRed}{RGB}{0,0,0}
\definecolor{MediumBlue}{RGB}{0,0,0}
\definecolor{Green}{RGB}{0,0,0}
\definecolor{DeepPink}{RGB}{0,0,0}
\definecolor{DarkOrange}{RGB}{0,0,0}

\usepackage{xpatch}
\makeatletter
\ExplSyntaxOn
\cs_new:Npn \bibColoredItems #1#2
  {
    \clist_map_inline:nn {#2} { \cs_new:cpn {bib@colored@##1} {#1} } 
  }
\ExplSyntaxOff

\newcommand\bib@setcolor[1]{%
  \ifcsname bib@colored@#1\endcsname
    \expandafter\color\expandafter{\csname bib@colored@#1\endcsname}
  \else
    \normalcolor
  \fi
}

\xpatchcmd\@bibitem
  {\item}
  {\bib@setcolor{#1}\item}
  {}{\PatchFailed}

\xpatchcmd\@lbibitem
  {\item}
  {\bib@setcolor{#2}\item}
  {}{\PatchFailed}
\makeatother
\def\BibTeX{{\rm B\kern-.05em{\sc i\kern-.025em b}\kern-.08em
    T\kern-.1667em\lower.7ex\hbox{E}\kern-.125emX}}
\begin{document}
% \nocite{*}
\title{AMLP: Adjustable Masking Lesion Patches for Self-Supervised Medical Image Segmentation}
\author{Xiangtao Wang, Ruizhi Wang, Thomas Lukasiewicz, Zhenghua Xu, \IEEEmembership{Member IEEE}
\thanks{This work was supported by the National Natural Science Foundation of China under the grant 62276089, by the Natural Science Foundation of Tianjin, China, under the grant 24JCJQJC00200, and by the Natural Science Foundation of Hebei Province, China, under the grant F2024202064. (Corresponding author: Zhenghua Xu, e-mail: zhenghua.xu@hebut.edu.cn.)}
\thanks{Xiangtao Wang, Ruizhi Wang, and Zhenghua Xu are with State Key Laboratory of Reliability and Intelligence of Electrical Equipment, School of Health Sciences and Biomedical Engineering, Hebei University of Technology, Tianjin, China.}
\thanks{Thomas Lukasiewicz is with Department of Computer Science, University of Oxford, Oxford, United Kingdom, and the institute of Logic and Computation, Vienna University of Technology, Vienna, Austria.}
}

% This work was supported by the National Natural Science Foundation of China under the grant 62276089, by the Natural Science Foundation of Tianjin, China, under the grants 24JCJQJC00200 and 24JCQNJC01230, by the Natural Science Foundation of Hebei Province, China, under the grant F2024202064, by the Science Research Project of Hebei Education Department, China, under the grant BJ2025004, by the Ministry of Human Resources and Social Security, China, under the grant RSTH-2023-135-1, and by the S\&T Program of Hebei under the grant 24464401D.
\maketitle

\begin{abstract}
Self-supervised masked image modeling (MIM) methods have shown promising performances on analyzing natural images. However, directly applying such methods to medical image segmentation tasks still cannot achieve satisfactory results. The challenges arise from the facts that (i) medical images are inherently more complex compared to natural images, and the subjects in medical images often exhibit more distinct contour features; (ii) moreover, the conventional high and fixed masking ratio in MIM is likely to mask the background, limiting the scope of learnable information. To address these problems, we propose a new self-supervised medical image segmentation framework, called Adjustable Masking Lesion Patches (AMLP), which employs Masked Patch Selection~(MPS) strategy to identify patches with high probabilities of containing lesions to help model achieve precise lesion reconstruction. To improve the categorization of patches in MPS, we further introduce Relative Reconstruction Loss (RRL) to better learn hard-to-reconstruct lesion patches. Then, Category Consistency Loss (CCL) is proposed to refine patch categorization based on reconstruction difficulty, enhancing difference between lesions and backgrounds. Moreover, an Adjustable Masking Ratio (AMR) strategy is proposed to gradually increase the masking ratio over training to expand~the scope of learnable mutual information. Extensive~experiments on two medical segmentation datasets demonstrate the superior performances of the proposed AMLP w.r.t. the SOTA self-supervised methods; the results prove that AMLP effectively addresses the challenges of applying masked modeling to medical images and capturing accurate lesion details that are crucial for segmentation tasks. 
\end{abstract}


\vspace{-0.5em}
\section{Introduction}
\label{sec:introduction}
\IEEEPARstart{S}{upervised} deep learning has achieved significant success in medical image segmentation tasks, leveraging a substantial amount of labeled data \cite{w-Net2022,kamnitsas2017efficient,mu-net2023}. However, a vast number of medical images are generated in routine clinical practice, but annotating them requires highly specialized skills, typically performed by radiologists with extensive clinical experience. Due to the limited number of specialized radiologists, their time constraints, and the annotation efficiency, acquiring large-scale medical image datasets with precise annotations is often very hard. This limitation hinders the use of supervised learning-based segmentation methods in routine clinical practice.
%%%  1 
Consequently, recent research has seen significant efforts in self-supervised medical image segmentation \cite{Jiaojiao-TMI2023}. Self-supervised learning aims to capture general features from unlabeled data and get the pre-trained model, and then transfer the pre-trained model into downstream tasks for fully supervised finetuning on a small amount of labeled data. Therefore, the performance of self-supervised medical image segmentation heavily relies on the quality of self-supervised pretraining.

\begin{figure}[!t]
	\centering
    \begin{minipage}{0.5\textwidth}
    \includegraphics[width=\textwidth]{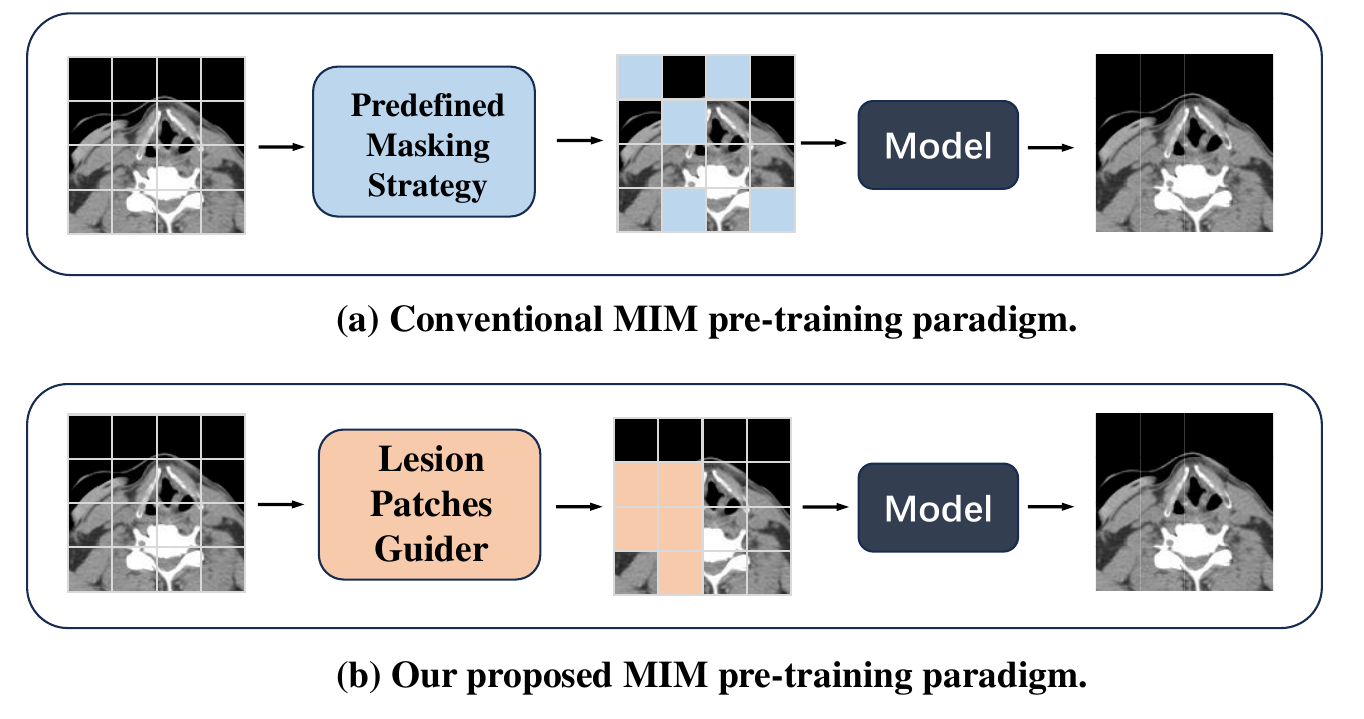}
    \end{minipage}
	\caption{Comparison between conventional MIM pretraining paradigm and proposed AMLP. (a) Conventional approaches solve a given problem under pre-defined mask strategy. (b) Our proposed AMLP guides model to mask lesion patches to produce a challenging pretext task.}
	\label{fig:littleFigure}
    \vspace{-2em}
\end{figure}

Classic self-supervised learning methods heavily rely on contrastive learning approaches. Contrastive learning (CL) aims to extract representative information by minimizing the distance between similar pairs of images while maximizing the distance between dissimilar pairs of images. However, these methods often prioritize capturing global semantics in images and may sometimes sacrifice local details and non-object regions \cite{chen2022context}. To address these challenges, researchers, inspired by masked language modeling (MLM) in natural language processing (NLP), have turned to masked image modeling (MIM) methods to explore potential solutions. Masked image modeling is designed to reconstruct masked regions within images, which typically contain information about important structures.

However, applying MIM methods used in natural images directly to medical images presents limitations. This is because there are notable differences between natural and medical images, including the small size and specific morphological features of lesions, which can lead models to unintentionally overlook or fully mask critical regions, and reduce downstream performance \cite{shen2017deep,litjens2017survey}. Furthermore, conventional MIM methods often employ fixed high masking ratios, which may hinder the effective capture of essential local information, emphasizing global image structures throughout encoding and decoding. This global-focused approach may result in neglecting crucial local details, e.g., the precise localization and morphology of lesions, which are indispensable in medical image analysis.

In this paper, to address these challenges, we introduce a novel self-supervised medical image segmentation method, named \textbf{A}daptive \textbf{M}asked \textbf{L}esion \textbf{P}atch (\textbf{AMLP}). Compared to existing methods, AMLP incorporates four key improvement measures: Masked Patch Selection (MPS), Relative Reconstruction Loss (RRL), Category Consistency Loss (CCL), and Adjustable Masking Ratio (AMR) strategies.

To solve the problem that conventional MIM methods are not suitable for medical images, we propose MPS. MPS can learn lesion representation by choosing the patches containing the lesion to mask. Specifically, we input all patches into k-means clustering, classify them into foreground and background, and then we rank the image patches in descending order based on the predicted probability of being foreground. Patches with higher foreground probabilities are masked first. This allows us to selectively prioritize masked image patches containing the lesion area, which can help learn lesion representation information more effectively.

However, the unsupervised k-means clustering applied in MPS may encounter the issue of potential misclassification. we further calculate the RRL for each patch, which is the ratio of the reconstruction loss to the average reconstruction loss, to ensure that the model accurately focuses on lesion patches during the training process. Additionally, we introduce the CCL, which helps the network distinguish differences between the reconstructed and original images in terms of categories, leading to a more accurate foreground-background distinction and enhancing the ability to learn lesion representation information.
We also propose the AMR strategy, which gradually increases the masking ratio and the final upper limit of the masking ratio, aiding the network in learning more global information in the initial encoding phases and raising the ability of learning representation information, ultimately enhancing the acquisition of more representation information and improving overall performance.

The contributions can be summarized as follows:
%\begin{itemize}[label={\textbullet}]
    %\item 
    (i) We identify the limitations under the applications of existing self-supervised MIM learning methods in medical images, and propose a novel adjustable masking lesions patches benefits self-supervised medical image segmentation framework, named AMLP.
    %\item 
    (ii) Our AMLP framework includes four key advancements. First, we introduce MPS based on k-means clustering to improve the model's focus on lesion patches. Second, the RRL calculates relative relative reconstruction weights to emphasize challenging lesion patches. Additionally, our CCL enhances the network's ability to distinguish between reconstructed and original images, improving foreground-background separation. Furthermore, we propose AMR to further enhance the model's performance by progressively increase masking ratio during training to broaden the range of learnable mutual information.  
    %\item 
    (iii)~Extensive experiments were conducted on two publicly available medical segmentation datasets. The results demonstrate that our method, with only 5\% or 10\% labeled data, AMLP greatly outperforms the state-of-the-art self-supervised baselines. Furthermore, ablation studies confirm the effectiveness and necessity of the four improvements (i.e., MPS, RRL, CCL, AMR) for AMLP to achieve its exceptional performance.
%\end{itemize}

\section{Related Work}
\label{sec:relatedwork}

In recent years, self-supervised learning in computer vision has garnered attention. Early methods focused on learning feature representations through data augmentation. However, these methods tend to prioritize global semantics over local details and non-object regions \cite{chen2022context}. A recent advancement named masked image modeling offers a fresh perspective on pretraining, which revolves around learning by reconstructing occluded image regions.

\begin{table}[!t]
    % \color{DeepPink}
    \vspace{0.5ex}
    \caption{Frequently Used Symbols.}
    \centering
    \resizebox{0.45\textwidth}{!}{
        \begin{tabular}{c|c}
            \toprule
            Symbol & Explanation                       \\
            \midrule
            $x$    & original image \\
            $x_m$    & masked image \\
            $x_i$    & masked patches \\
            $x_{\neg{i}}$    & unmasked patches \\
            $N$    & number of patches \\
            \midrule
            $L_{pred}$    & reconstruction loss \\
            $L_{mean}$    & mean reconstruction loss of all masked patches\\
            $L_{rrl}$    & relative reconstruction loss \\
            $L_{ccl}$    & category consistency loss\\
            \bottomrule
        \end{tabular}}
    \centering
    \label{tab:formla1}
    \vspace{-1ex}
\end{table}

\begin{figure*}[!t]
    \centering
    % \vspace{-2em}
    \includegraphics[width=0.95\textwidth]{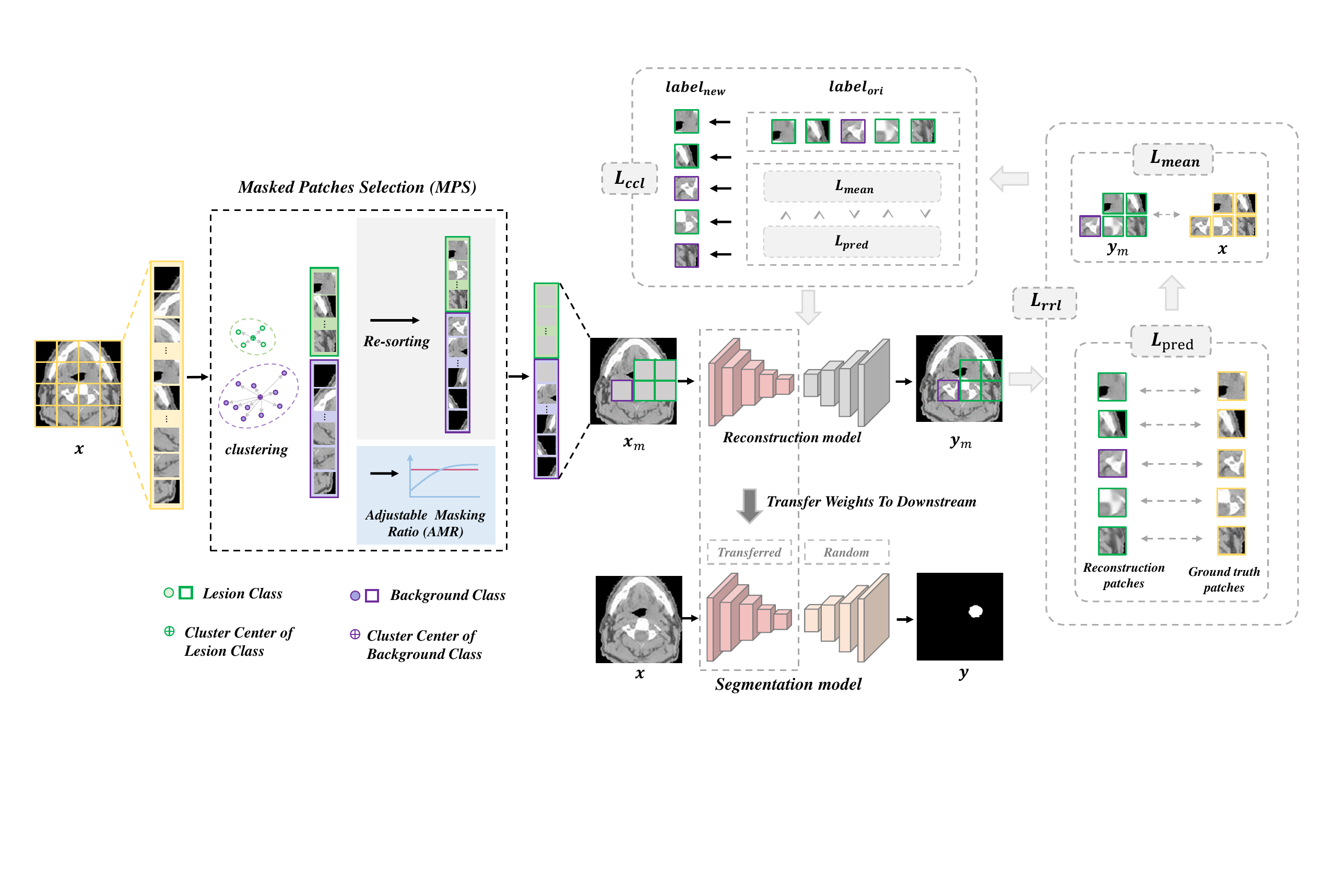}
    \vspace{-0.5em}
    \caption{Illustration of our proposed AMLP architecture. The image $x$ is divided into $N$ patches, which serve as the input of K-means clustering. The clustering process classifies the patches into two categories: lesion and background categories, and selectively mask them by the order of lesion-background. Then, the masked image is input into the reconstruction network for reconstruction. 
    After pretraining, only the encoder of reconstruction network is transferred to downstream segmentation model.}

    \vspace{-1.5em}
    \label{fig:main_network}
\end{figure*}

\vspace{-0.5em}
\subsection{Contrastive learning}
The objective of contrastive learning \cite{ZHANG-MIA2022} aims to learn representations by drawing positive samples closer and pushing negative samples further away in the latent space, thereby distinguishing each image from others. This goal is typically achieved by applying a contrastive loss. 
Currently, many contrastive learning methods have found successful applications in medical image segmentation. For instance, SimCLR \cite{chen2020simple} employs a contrastive loss function and multi-layer feature representation learning to facilitate self-learning on large-scale unlabeled data, thereby enhancing the model's representational and generalization capabilities, but this method relies too much on data augmentation. To mitigate this issue, MoCo \cite{he2020momentum}  further bolsters the model's robustness and discriminative performance by introducing momentum contrastive learning and a nonlinear contrastive loss function, expanding the set of negative samples and enhancing sample variability, but it only focuses on sample-level information and ignores details. Considering that possible negative pair construction errors could affect model performance, BYOL \cite{grill2020bootstrap} promotes the learning of more discriminative feature representations through bootstrapping methods and multi-task training, achieving significant improvements in SSL performance. To further process fine-grained features and complex scenes, SwAV \cite{caron2020unsupervised} proposes a method for unsupervised learning of visual features by contrastive cluster assignment, achieving effective feature and representation learning by comparing the similarity of samples in different cluster groupings.
However, more professional solutions are needed in the field of medical image analysis, G-L \cite{chaitanya2020contrastive} proposes to combine a local contrastive strategy (i.e., using 3D inter-layer location information for contrastive learning) with a global contrastive strategy to learn beneficial information representations.
To achieve deeper feature learning, PCRL \cite{zhou2021preservational} introduces Preservational Contrastive Representation Learning to reconstruct diverse contexts using representations learned from the contrastive loss.
To capture rich semantic information at both global and local levels, vox2vec \cite{goncharov2023vox2vec} models voxel representations through a Feature Pyramid Network (FPN), producing similar representations for the same voxels and unique representations for different voxels in various augmentation contexts, thereby learning multi-scale representations that capture global semantics and local semantics.
While these multi-scale representations are effective, they may not guarantee optimal alignment between features from different semantic levels. SAL \cite{Li2024SelfSupervisedAL} introduces a self-supervised alignment learning framework that enhances consistency in feature learning by aligning pixel-level features of matched structures across different semantic levels, thereby improving the robustness of medical image segmentation.
Besides the methods rely on invariance under certain intensity-based augmentations and geometric augmentations, Pix2Rep \cite{seince2024dense} proposes a pixel-level representation learning framework based on equivariance under geometric transformations. By explicitly modeling how pixel-level features should transform under different geometric operations, Pix2Rep enables more precise and geometrically-aware feature learning.

However, these methods often prioritize global semantic features over image details and non-subject regions~\cite{chen2022context}, affecting the performance in downstream segmentation tasks.

\vspace{-0.5em}
\subsection{Masked image modeling}
Masked image reconstruction involves masking specific image blocks and using a reconstruction network to recover them. For instance, MAE \cite{he2022masked} utilizes an asymmetric encoder-decoder structure, dividing the image into equal-sized blocks and predicting the masked block based on the unmasked block of the image. At the same time, SimMIM \cite{xie2022simmim} adjusts decoder weights and incorporates visible and masked patches as input, achieving results comparable to MAE. 
To address the imbalance problem of feature learning in different regions of the image, Maskfeat \cite{wei2022masked} introduces the Masked Feature Prediction method, enabling the model to predict masked features by randomly covering patches in the HOG feature map. ConvMAE \cite{gao2022convmae} enhances semantic information acquisition through MAE-based multi-scale encoding operations to learn more refined and discriminative feature representations. 
To further expand the application scope of masked image modeling, BEiT V2 \cite{peng2022beit} incorporates a discrete visual codebook generated by vector-quantized knowledge distillation (VQ-KD) and introduces a pre-training stage for high-level semantic abstraction.
Despite these advancements, model learning efficiency remains low. BootMAE \cite{dong2022bootstrapped} introduces a supervised Bootstrapped Masked Autoencoder, using masked autoencoding objectives for patch-level feature prediction tasks and an additional supervised bootstrapping signal. To optimize for the specificity of medical imaging, SelfMedMAE \cite{zhou2023self} proposes a self-pretraining paradigm with MAE for medical image analysis tasks, pretraining a ViT on the target data's training set instead of another dataset.

These methods solve specific problems under predefined masking strategies \cite{li2022uniform,kakogeorgiou2022hide}. Unlike existing methods, AMLP considers the unique importance of lesion areas in medical images, learns key representation information by specifically masking lesion areas, and achieves gradual feature learning through an adjustable masking ratio adjustment strategy. This allows AMLP to learn representations that are more beneficial to downstream segmentation tasks and achieve more accurate lesion segmentation.

\vspace{-0.5em}
\section{Methodology}
\label{sec:methodology}

Fig. \ref{fig:main_network} provides an overview of our proposed AMLP, which comprises four key components: Masked Patches Selection(MPS), Relative Reconstruction Loss (RRL), Category Consistency Loss (CCL), and Adjustable Masking Ratio (AMR). Specially, we commence by elucidating the masked patches selection strategy in Sec. \ref{sec:mps}. Furthermore, Sec. \ref{sec:l_arl} details the relative reconstruction loss, while Sec. \ref{sec:l_ccl} introduces the category consistency loss. Subsequently, the adjustable masking strategy is presented in Sec. \ref{sec:ams}.

\vspace{-1em}
\subsection{Masked Patches Selection}
\label{sec:mps}
Given that most current efforts in masked image modeling concentrate on natural images, with scant applications in medical images, we propose a masked patches selection strategy tailored specifically for medical images. 

Overall, the proposed masked patches selection (MPS) strategy is implemented by filtering and masking image patches containing lesions. First, the k-means algorithm is used for preliminary classification of the image patches, which are then prioritized based on their probability of containing lesions. Finally, the MPS selects and masks patches with higher probabilities to help the model specifically learn lesion features.

Before clustering begins, the input medical image $x$ is divided into $N$ equal-sized patches. Each patch is independently vectorized into a one-dimensional vector. Subsequently, the vectorized results of all image patches are concatenated into a two-dimensional array for clustering using k-means.

Since our ultimate goal is lesion segmentation, the k-means algorithm only needs to classify all image patches into two categories: foreground (i.e., patches containing lesion areas) and background (i.e., patches not containing lesions). Specifically, the k-means algorithm first randomly selects two patches as initial cluster centers. For each patch, calculate its distance to the two cluster centers and assign it to the category corresponding to the nearest one. Then, for each category, recalculate its cluster center (i.e., the centroid of all samples belonging to that category). These processes iterate continuously to optimize the clustering results.

We divide all patches into two categories $x=\lbrace x_{i}, x_{\neg{i}}\rbrace$ through k-means clustering. Considering the lesion areas are smaller than background areas, we set the group with the lesser number is the foreground $x_{i}$ and the bigger number is the background $x_{\neg{i}}$. They are assigned category labels of $1$ and $0$ respectively, e.g., $label_ {x_{i}}=1$, $label_{x_{\neg{i}}}=0$. Then, according to the probability of being judged as the foreground, they are sorted from large to small, and a foreground-background ordered mask is performed according to the mask ratio. Specially, the larger the distance from each patch to the center point of the foreground category, the greater the dissimilarity between its feature and the foreground cluster center, and accordingly, the greater the possibility that the patch is classified as foreground \cite{DHANACHANDRA2015764}.

\vspace{-1em}
\subsection{Relative Reconstruction Loss}
\label{sec:l_arl}
In the Masked Patch Selection (MPS), the sole reliance on unsupervised k-means clustering may lead to potential misclassification issues. Therefore, it is imperative to assist the network in accurately identifying foreground image patches. The most common method for determining whether a patch is foreground region is comparing the reconstruction loss $L_{pred}$ of a single patch, such as the $L_2$ loss used in MAE. (In this paper, we denote the $L_2$ loss as $L_{pred}$.) A higher loss value indicates a greater difficulty in reconstruction, which means it is more likely to be a foreground region. However, as training progresses, the scale of the $L_{pred}$ loss diminishes. When the loss values become very small, the model struggles to effectively differentiate the reconstruction difficulty of different patches, making it challenging to identify which patches need focused attention and masking. Therefore, the exclusive use of one patch's reconstruction loss is not the optimal choice.

To enhance the model's focus on the foreground regions more precisely, we propose a relative loss based on the Relative Reconstruction Loss (RRL). Unlike comparing the reconstruction loss of a single patch, the RRL dynamically evaluates the importance of each patch by calculating the ratio of a patch's reconstruction loss to the average reconstruction loss of all masked patches. When a patch's reconstruction loss is higher than the average, it indicates that the patch is more challenging in the reconstruction task, which usually means it is a foreground region. Consequently, RRL encourages the model to pay special attention to these regions, as they are more likely to contain important lesion information. Through this method, the model can more effectively identify and concentrate on foreground regions that require significant masking, thereby alleviating the issue of misclassification.

Specifically, we compute the relative reconstruction loss $L_{rrl}$ by calculating the ratio of reconstruction loss $L_{pred}$ for an patch to the average reconstruction loss of all masked image patches $L_{mean}$. The formula of $L_{rrl}$ is as follows:

\vspace{-0.5em}
\begin{small}
\begin{equation}
    L_{rrl}^{i} = {L_{pred}^{i}} / L_{mean},
\end{equation}
\end{small}

\noindent where $L_{pred}^{i}$ indicates the $L_{pred}$ loss of $i-th$ patch. And the formula of $L_{mean}$ is below: 

\vspace{-0.5em}
\begin{small}
\begin{equation}
    L_{mean} = {\sum\limits_{i=1}^N (L_{pred}^{i})}/N.
\end{equation}
\end{small}

\noindent where $N$ indicates the number of masked patches, and $L_{pred}^{i}$ indicates the $L_{pred}$ loss of $i-th$ patch.

\vspace{-0.5em}
\subsection{Category Consistency Loss}
\label{sec:l_ccl}
Furthermore, we propose the Category Consistency Loss (CCL) to enhance the network's ability to distinguish between foreground and background to further address the issue of misclassification. 
The CCL emphasizes patches whose classifications change during training by measuring the consistency between $Label_{new}$ and $Label_{ori}$.

Specifically, regarding the k-means clustering results, each patch is assigned a category label $Label_{ori}$ after k-means clustering, where the foreground category is represented by 1 ($Label_{ori} = 1$) and the background category is represented by 0 ($Label_{ori} = 0$). Concerning the relative values of $L_{pred}$ and $L_{mean}$, if $L_{pred} \geq L_{mean}$, the patch is considered more important compared to the average patch, and it is more likely to belong to the lesion category; if $L_{pred} < L_{mean}$, it is more likely to belong to the background category.

CCL updates the category labels of the patch clustering results according to the following rules: (i)If the k-means clustering result for a patch is consistent with the relative values of $L_{pred}$ and $L_{mean}$ (i.e., $Label_{ori} = 1$ and $L_{pred} \geq L_{mean}$, or $Label_{ori} = 0$ and $L_{pred} < L_{mean}$), the new category label for the patch remains unchanged; (ii)If the k-means clustering result for a patch is inconsistent with the relative values of $L_{pred}$ and $L_{mean}$ (i.e., $Label_{ori} = 1$ and $L_{pred} < L_{mean}$, or $Label_{ori} = 0$ and $L_{pred} \geq L_{mean}$), the new category label for the patch is randomly chosen from [0,1]. The reason for adopting the second update strategy is that it is difficult to determine which is more accurate between the k-means clustering result and the relative values of $L_{pred}$ and $L_{mean}$. 
Therefore, to balance these two situations, we use a random selection method.

\begin{figure}[!t]
    \centering
    \begin{minipage}{0.48\textwidth}
    \includegraphics[width=\textwidth]{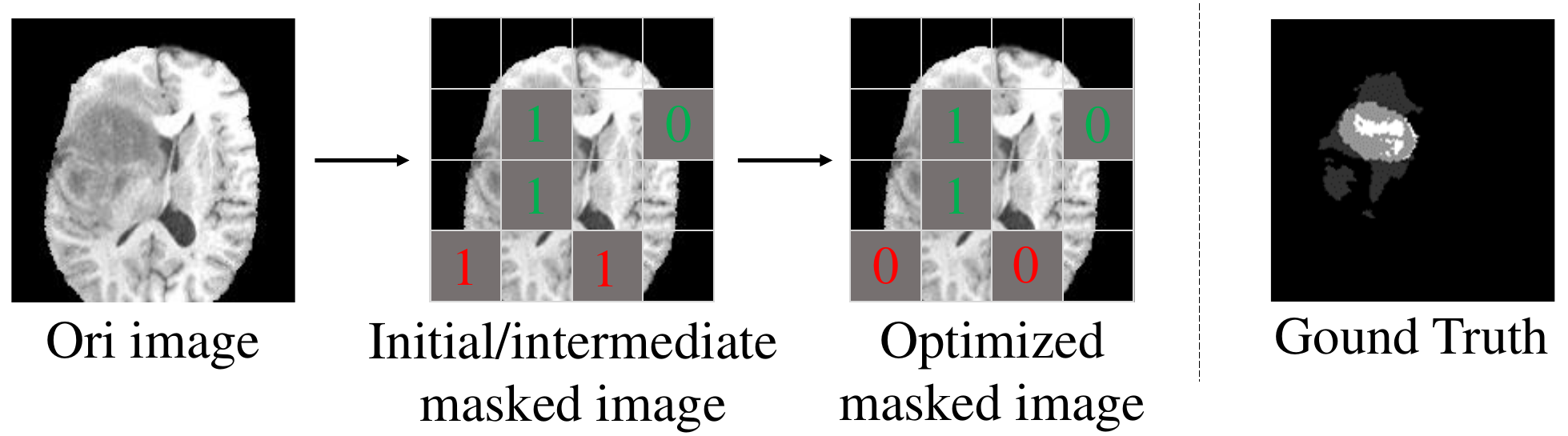}
    \vspace{-1.5em}
    \end{minipage}
	\caption{Visualization of the patch category correction process during training, where 1 represents the lesion category, and 0 represents the background category, and green represents correctly classified image patches, and red represents incorrectly classified image patches.}
	\label{fig:labelchange}
    \vspace{-1.5em}
\end{figure}

To facilitate a more intuitive understanding of this process, we have visualized the patch category correction during training in Figure \ref{fig:labelchange}. 
Correct and incorrect classifications (compared to the ground truth) are highlighted in green and red, respectively. As training progresses, the two misclassified foreground red image patches (label 1) are corrected to background (label 0), indicating that our proposed AMLP can dynamically adjust and improve the classification results of image patches during the model learning process.

When $label_{new} \neq label_{ori}$, indicating a change in patch categorization during training, we calculate CCL using the following formula: 

\vspace{-0.5em}
\begin{small}
\begin{equation}
    \begin{aligned} 
        L_{ccl} = - [ label_{new} \times \log(label_{ori} + \xi) \\
         +  (1 - label_{new}) \times \log(1 - label_{ori} + \xi)].
         \vspace{-0.5em}
    \end{aligned}
\end{equation}
\end{small}

where $Label_{ori}$ and $Label_{new}$ can only take the values 0 and 1, and $\xi = 1e-1$ is used to prevent $\log(0)$. After each training epoch, the $label_{new}$ will replace the current $label_{ori}$ for the corresponding patch to enable dynamic label updating.
Furthermore, we observed that the final results converge to the first update rule.

The overall loss of AMLP is defined as:

\vspace{-0.5em}
\begin{small}
\begin{equation}
    L_{all} = L_{pred} + L_{rrl} + L_{ccl}.
    \vspace{-0.5em}
\end{equation}
\end{small}

\vspace{-0.5em}
\subsection{Adjustable Masking Ratio}
\label{sec:ams}

Given that a higher mask ratio can force the model to learn more challenging and rich features. We question whether the commonly used fixed masking ratio of 75\% in studies like MAE \cite{he2022masked} and SimMIM \cite{xie2022simmim} is the upper limit. Could we further improve the representation learning ability and achieve better performance by increasing the upper limit of the masking ratio?

To overcome this limitation, we propose an innovative dynamic masking ratio strategy, where the masking ratio gradually increases from an initial value as training progresses. The initial masking ratio $\sigma_0$ is set to $25\%$, based on experimental findings in MAE \cite{he2022masked}. Furthermore, as training progresses, the masking ratio increases according to the following equation and finally reaches 80\%:

\vspace{-0.5em}
\begin{small}
\begin{equation}
    \sigma = \sigma_0 + {\ln(x_e)}/{\tau},
    \vspace{-0.5em}
\end{equation}
\end{small}

\noindent where $x_e$ denotes the training epoch, and $\tau$ is a manually defined constant.

\begin{table*}[!t] 
    \begin{center}
    \vspace{-1em}
        \caption{The DSC and Sen results of AMLP and the state-of-the-art fully-supervised and self-supervised segmentation methods on the Hecktor and BraTS2018 datasets with 5\%, and 10\% labeled data in terms of DSC and Sen.}
        \vspace{-0.5em}
        \label{table:main_1}
        \setlength\tabcolsep{2pt}
        \newcolumntype{"}{@{\hskip\tabcolsep\vrule width 1pt\hskip\tabcolsep}}
        \begin{tabular}{m{0.04\textwidth}<{\centering}m{0.12\textwidth}<{\centering}m{0.06\textwidth}<{\centering}p{0.06\textwidth}<{\centering}p{0.06\textwidth}<{\centering}p{0.06\textwidth}<{\centering}p{0.06\textwidth}<{\centering}p{0.06\textwidth}<{\centering}p{0.06\textwidth}<{\centering}p{0.06\textwidth}<{\centering}p{0.06\textwidth}<{\centering}p{0.06\textwidth}<{\centering}p{0.06\textwidth}<{\centering}p{0.06\textwidth}<{\centering}}
            \hlineB{3}
            \multicolumn{2}{c}{\textbf{Methods}}
            & \multicolumn{4}{c}{\textbf{Hecktor}}     
            & \multicolumn{8}{c}{\textbf{BraTS2018}}  \\    
            % \multicolumn{2}{c}{}&\multicolumn{4}{c}{}&\multicolumn{8}{c}{}\\ 
            \cmidrule(lr){3-6} \cmidrule(lr){7-14}
            
            \multicolumn{2}{c}{Modality} &\multicolumn{2}{c}{CT} & \multicolumn{2}{c}{PET}& \multicolumn{2}{c}{T1CE}& \multicolumn{2}{c}{T2} 
            & \multicolumn{2}{c}{FLAIR} & \multicolumn{2}{c}{T1}   \\ \cmidrule(lr){3-4} \cmidrule(lr){5-6} \cmidrule(lr){7-8} \cmidrule(lr){9-10} \cmidrule(lr){11-12} \cmidrule(lr){13-14}
            
            \multicolumn{2}{c}{Metrics} 
            &{DSC$\uparrow$}&{Sen$\uparrow$}  &DSC$\uparrow$&{Sen$\uparrow$}  &DSC$\uparrow$&{Sen$\uparrow$}  &DSC$\uparrow$&{Sen$\uparrow$}   &DSC$\uparrow$&{Sen$\uparrow$}  
            &DSC$\uparrow$&\multicolumn{1}{c}{Sen$\uparrow$}\\ \cline{1-14}
            
            \multirow{15}{*}{\textbf{5\%}}{} &{Sup} 
            &{0.1740}&{0.2337} 
            &{0.5452}&{0.7044}
            &{0.3234}&{0.2703} 
            &{0.3952}&{0.4371}
            &{0.4035}&{0.4837}  
            &{0.1743}&{0.2702}
            \\ 
            
            \multicolumn{1}{c}{}&{SimCLR \cite{chen2020simple}} 
            &{0.1733}&{0.2527} 
            &{0.5720}&{0.7143}
            &{0.5403}&{0.5737} 
            &{0.4310}&{0.5870}
            &{0.4342}&{0.5396}  
            &{0.2908}&{0.4376} 
            \\ 
            
            \multicolumn{1}{c}{}&{BYOL \cite{grill2020bootstrap}}  
            &{0.1967}&{0.2555} 
            &{0.4487}&{0.5603}
            &{0.5245}&{0.4321} 
            &{0.4377}&{0.5590}
            &{0.4561}&{0.6297}  
            &{0.2811}&{0.4545} 
            \\ 
            
            \multicolumn{1}{c}{}&{SwAV \cite{caron2020unsupervised}}
            &{0.2307}&{0.2483} 
            &{0.5517}&{0.7091}
            &{0.5252}&{0.4321} 
            &{0.4160}&{0.5564}
            &{0.4346}&{0.6052} 
            &{0.2277}&{0.4466} 
            \\ %\hline

            \multicolumn{1}{c}{}&{G-L \cite{chaitanya2020contrastive}}
            &{0.1754}&{0.1719}
            &{0.5615}&{0.6993}
            &{0.3195}&{0.4130}
            &{0.4412}&{0.5281}
            &{0.4355}&{0.5603}
            &{0.2137}&{0.2917}
            \\ %\hline

            \multicolumn{1}{c}{}&{PCRL \cite{zhou2021preservational}} 
            &{0.2374}&{0.2777}
            &{0.5605}&{0.7130}
            &{0.5158}&{0.4572}
            &{0.4251}&{0.5227}
            &{0.4453}&{0.6124}
            &{0.2695}&{0.4365}
            \\ %\hline
            
            \multicolumn{1}{c}{}&{MAE \cite{he2022masked}} 
            &{0.2560}&{0.2966} 
            &{0.5931}&{0.7301}
            &{0.5465}&{0.5276} 
            &{0.4395}&{0.5134}
            &{0.4368}&{0.6108}  
            &{0.2734}&{0.4596} 
            \\ %\hline
            
            \multicolumn{1}{c}{}&{MaskFeat \cite{wei2022masked}} 
            &{0.2561}&{0.2894} 
            &{0.5900}&{0.7277}
            &{0.5332}&{0.5173} 
            &{0.4492}&{0.5012}
            &{0.4442}&{0.5937}  
            &{0.2834}&{0.4542} 
            \\ %\hline

            \multicolumn{1}{c}{}&{SimMIM \cite{xie2022simmim}} 
            &{0.2413}&{0.2972} 
            &{0.5892}&{0.7239}
            &{0.5421}&{0.5012} 
            &{0.4289}&{0.4901}
            &{0.4525}&{0.6013}  
            &{0.2801}&{0.4297} 
            \\ %\hline
            
            \multicolumn{1}{c}{}&{ConvMAE \cite{gao2022convmae}} 
            &{0.2611}&{0.3042} 
            &{0.6042}&{0.7431}
            &{0.5552}&{0.5414} 
            &{0.4497}&{0.5338}
            &{0.4247}&{0.6243} 
            &{0.2979}&{0.4865}
            \\ %\hline
            
            \multicolumn{1}{c}{}&{BEiT V2 \cite{peng2022beit}} 
            &{0.2759}&{0.3153} 
            &{0.6133}&{0.7548}
            &{0.5508}&\textbf{0.5779} 
            &{0.4504}&{0.5587}
            &{0.4501}&\textbf{0.6299} 
            &{0.3027}&{0.4891} 
            \\ %\hline
            
            \multicolumn{1}{c}{}&{BootMAE \cite{dong2022bootstrapped}} 
            &{0.2593}&{0.2978} 
            &{0.5988}&{0.7345}
            &{0.5431}&{0.5358} 
            &{0.4496}&{0.5134}
            &{0.4412}&{0.6164} 
            &{0.2908}&{0.4739} 
            \\ %\hline
            
            \multicolumn{1}{c}{}&{SelfMedMAE \cite{zhou2023self}}
            &{0.2699}&{0.3115} 
            &{0.6235}&{0.7439} 
            &{0.5523}&{0.5516} 
            &{0.4557}&{0.5284} 
            &{0.4594}&{0.6104} 
            &{0.2944}&{0.4751} 
            \\ %\hline

            \multicolumn{1}{c}{}&{vox2vec \cite{goncharov2023vox2vec}}
            &{0.2791}&{0.3193} 
            &{0.6240}&{0.7541} 
            &{0.5632}&{0.5635} 
            &{0.4605}&\textbf{0.5621} 
            &{0.4623}&{0.6229} 
            &{0.3011}&{0.4801} 
            \\ \cline{2-14}

            \multicolumn{1}{c}{} &{Ours} 
            &\textbf{0.2905}&\textbf{0.3347}   
            &\textbf{0.6296}&\textbf{0.7665}       
            &\textbf{0.5762}&0.5482 
            &\textbf{0.4721} &{0.5023}  
            &\textbf{0.4693} &{0.6136} 
            &\textbf{0.3297} &\textbf{0.5023} 
            \\  \cline{1-14}
            
            \multirow{15}{*}{\textbf{10\%}}{}&{Sup}
            &{0.2541}&{0.2703} 
            &{0.5769}&{0.7067}
            &{0.4451}&{0.3696} 
            &{0.4288}&{0.4716}
            &{0.4489}&{0.6225}  
            &{0.2476}&{0.3366} 
            \\ %\hline
            
            \multicolumn{1}{c}{}&{SimCLR \cite{chen2020simple}} 
            &{0.2951}&{0.3895} 
            &{0.5779}&{0.7145}
            &{0.6024}&{0.6217} 
            &{0.4846}&{0.5729}
            &{0.4720}&{0.5217}  
            &{0.3551}&{0.4868} 
            \\ %\hline
            
            \multicolumn{1}{c}{}&{BYOL \cite{grill2020bootstrap}}  
            &{0.3088}&{0.3930} 
            &{0.5891}&{0.6990}
            &{0.5991}&{0.6323} 
            &{0.4850}&{0.5710}
            &{0.4763}&{0.6329}  
            &{0.3458}&{0.3535} 
            \\ %\hline
            
            \multicolumn{1}{c}{}&{SwAV \cite{caron2020unsupervised}}
            &{0.2690}&{0.3340} 
            &{0.5771}&{0.7004}
            &{0.5515}&{0.5646} 
            &{0.4587}&{0.5794}
            &{0.4543}&{0.6084}  
            &{0.2914}&{0.4694} 
            \\ %\hline

            \multicolumn{1}{c}{}&{G-L \cite{chaitanya2020contrastive}} 
            &{0.2784}&{0.2945}
            &{0.5923}&{0.7304}
            &{0.4575}&{0.5045}
            &{0.4645}&{0.5718}
            &{0.4504}&\textbf{0.6691}
            &{0.2685}&{0.4086}
            \\ %\hline

            \multicolumn{1}{c}{}&{PCRL \cite{zhou2021preservational}}
            &{0.2907}&{0.3985} 
            &{0.5904}&{0.7039} 
            &{0.5649}&{0.5713} 
            &{0.4726}&{0.5672} 
            &{0.4601}&{0.6453} 
            &{0.3083}&{0.4461} 
            \\ %\hline
            
            \multicolumn{1}{c}{}&{MAE \cite{he2022masked}} 
            &{0.3195}&{0.4143} 
            &{0.6058}&{0.7155}
            &{0.5881}&{0.5947} 
            &{0.4765}&{0.5256}
            &{0.4587}&{0.6178}  
            &{0.3578}&{0.4878} 
            \\ %\hline
            
            \multicolumn{1}{c}{}&{MaskFeat \cite{wei2022masked}} 
            &{0.3078}&{0.4077} 
            &{0.5955}&{0.7132}
            &{0.6065}&{0.5864} 
            &{0.4613}&{0.5103}
            &{0.4625}&{0.6154}  
            &{0.3351}&{0.4800} 
            \\ %\hline

            \multicolumn{1}{c}{}&{SimMIM \cite{xie2022simmim}} 
            &{0.2920}&{0.4025} 
            &{0.5872}&{0.7221}
            &{0.5932}&{0.5701} 
            &{0.4592}&{0.5082}
            &{0.4572}&{0.5921}  
            &{0.3246}&{0.4725} 
            \\ %\hline
            
            \multicolumn{1}{c}{}&{ConvMAE \cite{gao2022convmae}} 
            &{0.3369}&{0.4273} 
            &{0.6099}&{0.7183}
            &{0.6106}&{0.6208} 
            &{0.5031}&{0.5667}
            &{0.4915}&{0.6246} 
            &{0.3617}&{0.5123} 
            \\ %\hline     
            
            \multicolumn{1}{c}{}&{BEIT V2 \cite{peng2022beit}} 
            &{0.3470}&{0.4388} 
            &{0.6293}&{0.7217}
            &{0.6139}&{0.6315} 
            &{0.5079}&{0.5897}
            &{0.4862}&{0.6446} 
            &{0.3690}&{0.5334} 
            \\ %\hline
            
            \multicolumn{1}{c}{}&{BootMAE \cite{dong2022bootstrapped}} 
            &{0.3225}&{0.4243} 
            &{0.6221}&{0.7155}
            &{0.6203}&{0.6032} 
            &{0.4857}&{0.5321}
            &{0.4751}&{0.6093} 
            &{0.3624}&{0.5011} 
            \\ %\hline
            
            \multicolumn{1}{c}{}&{SelfMedMAE \cite{zhou2023self}}
            &{0.3403}&{0.4313}
            &{0.6121}&{0.7204}
            &{0.6177}&{0.6159}
            &{0.5019}&{0.5330}
            &{0.4825}&{0.6264}
            &{0.3740}&{0.5322}
            \\ %\hline

            \multicolumn{1}{c}{}&{vox2vec \cite{goncharov2023vox2vec}}
            &{0.3509}&{0.4420}
            &{0.6285}&{0.7213}
            &{0.6268}&\textbf{0.6379}
            &{0.5107}&\textbf{0.5975}
            &{0.4904}&{0.6313}
            &{0.3798}&{0.5461}
            \\ \cline{2-14}
            
            \multicolumn{1}{c}{}&{Ours} 
            &\textbf{0.3690}&\textbf{0.5004}   
            &\textbf{0.6314}&\textbf{0.7331}       
            &\textbf{0.6343}&{0.6023}    
            &\textbf{0.5194}&{0.5482} 
            &\textbf{0.4971}&{0.6324} 
            &\textbf{0.3815}&\textbf{0.5782} 
            \\  \cline{1-14}

            \multicolumn{2}{c}{Sup (50\%)}
            &{0.3440}&{0.4331} 
            &{0.6252}&{0.6582}
            &{0.5941}&{0.5324} 
            &{0.5094}&{0.5328}
            &{0.4830}&{0.5664}  
            &{0.3581}&{0.4564} 
            \\ 

            \multicolumn{2}{c}{Sup (100\%)}
            &{0.3927}&{0.4736} 
            &{0.6475}&{0.7307}
            &{0.7453}&{0.7204} 
            &{0.5556}&{0.5575}
            &{0.5164}&{0.6561} 
            &{0.4207}&{0.4923} 
            \\  \hlineB{3} 	
            
        \end{tabular}
    \end{center}
    \vspace{-2em}
\end{table*}

By integrating the masked patches selection strategy with adjustable masking ratio strategy, we can compute the number of masked patches $n_m$ and the masked image $x_m$, where $n_m = \lfloor N \times \sigma \rfloor$. Then, the masked image $x_m$ is input into the reconstruction network to reconstruct the masked patches.

It is worth noting that AMLP uses the k-means clustering algorithm to select masked patches, introducing a certain degree of randomness. Although we later incorporate RRL and CCL to mitigate the inherent misclassification issues of k-means clustering, the lower masking rate in the early stages of training, combined with the randomness of the k-means algorithm in selecting lesion patches, ensures that the model can learn discriminative features from unmasked lesion patches. Therefore, even under a high masking ratio later on, where lesions are completely masked, the model can still perform lesion recognition and segmentation tasks based on its learned high representational capacity.

\vspace{-0.5em}
\section{Experiments and Results}
\label{sec:results}

\noindent\textbf{Datasets}:
To validate the effectiveness of our proposed AMLP, we conduct extensive evaluations on two publicly available medical image datasets, Hecktor~\cite{Hecktor1} and BraTS2018 ~\cite{Brats2018_1}. 
(\romannumeral1) \emph{Hecktor.} The dataset comes from the MICCAI $2020$ challenge called HEad and the neCK TumOR Segmentation Challenge (Hecktor) for Head and Neck Tumor Segmentation. It contains 201 3D head and neck PET-CT scans.
(\romannumeral2) \emph{BraTS2018.} This dataset was released by the BraTS'18 challenge hosted by MICCAI 2018 for brain tumor segmentation, including three hierarchical labels:  WT (whole tumor), ET (enhancing tumor), and TC (tumor core). It contains four modalities: T1, T1CE, T2 and FLAIR. 

% 分指标

% \begin{small}
\begin{table*}[!t] 
    % 	\vspace{-2em}
    % 		\footnotesize
    \begin{center}
        \caption{The BIoU and HD95 results of AMLP and the state-of-the-art fully-supervised and self-supervised segmentation methods on the Hecktor and BraTS2018 datasets with 5\%, and 10\% labeled data in terms of BIoU and HD95.
        }%
        \vspace{-0.5em}
        \smallskip 
        \label{table:main_2}
        \setlength\tabcolsep{2pt}
        \newcolumntype{"}{@{\hskip\tabcolsep\vrule width 1pt\hskip\tabcolsep}}
        \begin{tabular}{m{0.04\textwidth}<{\centering}m{0.12\textwidth}<{\centering}m{0.06\textwidth}<{\centering}p{0.06\textwidth}<{\centering}p{0.06\textwidth}<{\centering}p{0.06\textwidth}<{\centering}p{0.06\textwidth}<{\centering}p{0.06\textwidth}<{\centering}p{0.06\textwidth}<{\centering}p{0.06\textwidth}<{\centering}p{0.06\textwidth}<{\centering}p{0.06\textwidth}<{\centering}p{0.06\textwidth}<{\centering}p{0.06\textwidth}<{\centering}}
            \hlineB{3}
            \multicolumn{2}{c}{\textbf{Methods}}
            & \multicolumn{4}{c}{\textbf{Hecktor}}     
            & \multicolumn{8}{c}{\textbf{BraTs2018}} \\ 
            % \multicolumn{2}{c}{}&\multicolumn{4}{c}{}&\multicolumn{8}{c}{}\\ 
            \cmidrule(lr){3-6} \cmidrule(lr){7-14}
            
            \multicolumn{2}{c}{Modality} &\multicolumn{2}{c}{CT} & \multicolumn{2}{c}{PET}& \multicolumn{2}{c}{T1CE}& \multicolumn{2}{c}{T2} 
            & \multicolumn{2}{c}{FLAIR} & \multicolumn{2}{c}{T1}   \\ \cmidrule(lr){3-4} \cmidrule(lr){5-6} \cmidrule(lr){7-8} \cmidrule(lr){9-10} \cmidrule(lr){11-12} \cmidrule(lr){13-14}
            
            \multicolumn{2}{c}{Metrics} 
            &{BIoU$\uparrow$}&{HD95$\downarrow$}  &{BIoU$\uparrow$}&{HD95$\downarrow$}   &{BIoU$\uparrow$}&{HD95$\downarrow$}   &{BIoU$\uparrow$}&{HD95$\downarrow$}    &{BIoU$\uparrow$}&{HD95$\downarrow$}  &BIoU$\uparrow$&\multicolumn{1}{c}{HD95$\downarrow$}\\ 
            \cline{1-14}

            \multirow{15}{*}{\textbf{5\%}}{} &{Sup} 
            &{0.0690}&{62.7115} 
            &{0.2493}&{69.0963}
            &{0.1487}&{55.0727} 
            &{0.1330}&{47.4584}
            &{0.2055}&{33.2086}  
            &{0.0484}&{59.5963} 
            \\ %\hline

            \multicolumn{1}{c}{}&{SimCLR \cite{chen2020simple}} 
            &{0.0718}&{61.0090} 
            &{0.2627}&{64.9785}
            &{0.1492}&{54.0091} 
            &{0.1357}&{46.3305}
            &{0.2072}&{32.7433}  
            &{0.0567}&{57.6039}
            \\ %\hline
            
            \multicolumn{1}{c}{}&{BYOL \cite{grill2020bootstrap}}  
            &{0.0845}&{60.8153} 
            &{0.2793}&{60.2143}
            &{0.1504}&{49.3725} 
            &{0.1501}&{44.2871}
            &{0.2103}&{31.0754} 
            &{0.0609}&{54.4291} 
            \\ %\hline
            
            \multicolumn{1}{c}{}&{SwAV \cite{caron2020unsupervised}}
            &{0.0711}&{61.9957} 
            &{0.2678}&{61.8257}
            &{0.1495}&{52.2436} 
            &{0.1401}&{45.8792}
            &{0.2083}&{31.5867}  
            &{0.0589}&{56.4368}
            \\ %\hline

            \multicolumn{1}{c}{}&{G-L \cite{chaitanya2020contrastive}} 
            &{0.0703}&{62.2405} 
            &{0.2561}&{62.0235} 
            &{0.1506}&{52.0623} 
            &{0.1463}&{45.9033} 
            &{0.2100}&{32.0056} 
            &{0.0628}&{57.2391} 
            \\ %\hline

            \multicolumn{1}{c}{}&{PCRL \cite{zhou2021preservational}} 
            &{0.0904}&{60.1120} 
            &{0.2743}&{62.6295} 
            &{0.1562}&{53.2362} 
            &{0.1533}&{46.0758} 
            &{0.2154}&{31.9603} 
            &{0.0701}&{58.0055} 
            \\ %\hline
            
            \multicolumn{1}{c}{}&{MAE \cite{he2022masked}} 
            &{0.0999}&{58.7935} 
            &{0.2802}&{54.9321}
            &{0.1472}&{48.2318} 
            &{0.1489}&{43.1123}
            &{0.2106}&{30.2385} 
            &{0.0842}&{53.5203} 
            \\ %\hline
            
            \multicolumn{1}{c}{}&{MaskFeat \cite{wei2022masked}} 
            &{0.1067}&{59.0087} 
            &{0.2799}&{58.9967}
            &{0.1505}&{49.0699} 
            &{0.1463}&{43.7748}
            &{0.2117}&{32.7451}
            &{0.0775}&{54.2226} 
            \\ %\hline

            \multicolumn{1}{c}{}&{SimMIM \cite{xie2022simmim}} 
            &{0.0793}&{60.6234} 
            &{0.2743}&{58.4102}
            &{0.1461}&{51.0855} 
            &{0.1382}&{44.2219}
            &{0.2095}&{31.1998} 
            &{0.0589}&{55.0105} 
            \\ %\hline
            
            \multicolumn{1}{c}{}&{ConvMAE \cite{gao2022convmae}} 
            &{0.1053}&{58.5928} 
            &{0.2955}&{50.6654}
            &{0.1593}&{46.8451} 
            &{0.1574}&{41.0145}
            &{0.2140}&{29.1014} 
            &{0.1063}&{51.1002} 
            \\ %\hline
            
            \multicolumn{1}{c}{}&{BEiT V2 \cite{peng2022beit}} 
            &{0.1374}&{57.4124} 
            &{0.2994}&{48.2849}
            &{0.1607}&{46.3542} 
            &{0.1623}&f{40.3721}
            &{0.2152}&{28.5621}  
            &{0.1112}&{50.2631} 
            \\ %\hline
            
            \multicolumn{1}{c}{}&{BootMAE \cite{dong2022bootstrapped}} 
            &{0.1201}&{59.5928} 
            &{0.2836}&{52.7976}
            &{0.1512}&{47.1273} 
            &{0.1537}&{41.2457}
            &{0.2128}&{29.8902} 
            &{0.0928}&{52.4317} 
            \\ %\hline
            
            \multicolumn{1}{c}{}&{SelfMedMAE \cite{zhou2023self}}
            &{0.1414}&{59.7325}
            &{0.2910}&{51.1758}
            &{0.1672}&{46.1121}
            &{0.1608}&{40.1080}
            &{0.2104}&{29.7491}
            &{0.1074}&{51.4210}
            \\ %\hline

            \multicolumn{1}{c}{}&{vox2vec \cite{goncharov2023vox2vec}}
            &{0.1455}&{58.3407}
            &{0.3017}&{47.3999}
            &{0.1692}&{45.8761}
            &{0.1638}&\textbf{40.0012}
            &{0.2139}&{28.9602}
            &{0.1130}&{50.4658}
            \\ \cline{2-14}

            \multicolumn{1}{c}{} &{Ours} 
            &\textbf{0.1514}&\textbf{57.2823}   
            &\textbf{0.3113}&\textbf{40.5376}       
            &\textbf{0.1723}&\textbf{44.2326}    
            &\textbf{0.1750} &{40.6362} 
            &\textbf{0.2156} &\textbf{28.2352} 
            &\textbf{0.1157} &\textbf{50.2342} 
            \\ \cline{1-14}
            
            \multirow{15}{*}{\textbf{10\%}}{}&{Sup}
            &{0.1224}&{64.0604}
            &{0.3199}&{41.8627} 
            &{0.1643}&{50.9107} 
            &{0.1510}&{43.9483}
            &{0.1724}&{30.4918} 
            &{0.1042}&{52.8203} 
            \\ %\hline
            
            \multicolumn{1}{c}{}&{SimCLR \cite{chen2020simple}} 
            &{0.1504}&{46.3870} 
            &{0.3581}&{66.4539}
            &{0.2758}&{37.6094} 
            &{0.2020}&{38.0898}
            &{0.2203}&{28.5869}  
            &{0.1247}&{46.2637}
            \\ %\hline
            
            \multicolumn{1}{c}{}&{BYOL \cite{grill2020bootstrap}}  
            &{0.1501}&{58.9947} 
            &{0.3397}&{55.0181}
            &{0.2901}&{36.1710} 
            &{0.1820}&{42.4852}
            &{0.2075}&{27.9066} 
            &{0.1201}&{39.8026} 
            \\ %\hline
            
            \multicolumn{1}{c}{}&{SwAV \cite{caron2020unsupervised}}
            &{0.1543}&{59.5009} 
            &{0.3573}&{50.8417}
            &{0.2287}&{46.0846} 
            &{0.1889}&{44.0658}
            &{0.2036}&{38.6023} 
            &{0.1035}&{61.8490} 
            \\ %\hline

            \multicolumn{1}{c}{}&{G-L \cite{chaitanya2020contrastive}}
            &{0.1462}&{57.9235}
            &{0.3473}&{53.8722}
            &{0.2490}&{48.6177}
            &{0.1803}&{41.9554}
            &{0.2048}&{36.1221}
            &{0.0969}&{53.0828}
            \\ %\hline

            \multicolumn{1}{c}{}&{PCRL \cite{zhou2021preservational}} 
            &{0.1593}&{57.1364}
            &{0.3603}&{50.1351}
            &{0.2587}&{40.8316}
            &{0.1974}&{41.0274}
            &{0.2109}&{35.1126}
            &{0.1197}&{50.1286}
            \\ %\hline
            
            \multicolumn{1}{c}{}&{MAE \cite{he2022masked}} 
            &{0.1702}&{55.2376} 
            &{0.3445}&{46.1427}
            &{0.3113}&{32.7709} 
            &{0.2167}&{32.9846}
            &{0.2145}&{26.9349} 
            &{0.1182}&{35.9123} 
            \\ %\hline
            
            \multicolumn{1}{c}{}&{MaskFeat \cite{wei2022masked}} 
            &{0.1634}&{57.2447} 
            &{0.3618}&{47.6183}
            &{0.3024}&{33.2451} 
            &{0.1987}&{35.6779}
            &{0.2109}&{27.3426}
            &{0.1245}&{37.5428} 
            \\ %\hline

            \multicolumn{1}{c}{}&{SimMIM \cite{xie2022simmim}} 
            &{0.1601}&{58.3412} 
            &{0.3511}&{48.8915}
            &{0.2975}&{34.8762} 
            &{0.2063}&{37.5124}
            &{0.2096}&{29.8875}  
            &{0.1223}&{39.3501} 
            \\ %\hline
            
            \multicolumn{1}{c}{}&{ConvMAE \cite{gao2022convmae}} 
            &{0.1882}&{50.3523} 
            &{0.3739}&{41.7655}
            &{0.3421}&{25.9865} 
            &{0.2301}&{28.0147}
            &{0.2012}&{25.6731}  
            &{0.1356}&{27.8754} 
            \\ %\hline     
            
            \multicolumn{1}{c}{}&{BEIT V2 \cite{peng2022beit}} 
            &{0.1990}&{48.7192} 
            &{0.3899}&{39.8887}
            &{0.3578}&{22.4567} 
            &{0.2253}&{23.8829}
            &{0.2218}&{24.2587} 
            &{0.1428}&{24.7159} 
            \\ %\hline
            
            \multicolumn{1}{c}{}&{BootMAE \cite{dong2022bootstrapped}} 
            &{0.1788}&{54.5214} 
            &{0.3612}&{42.9639}
            &{0.3182}&{27.9943} 
            &{0.2173}&{30.5402}
            &{0.2186}&{25.9823} 
            &{0.1324}&{33.7837} 
            \\ %\hline
            
            \multicolumn{1}{c}{}&{SelfMedMAE \cite{zhou2023self}}
            &{0.1960}&{46.6575}
            &{0.3933}&{40.1080}
            &{0.3535}&{23.6667}
            &{0.2105}&{26.2442}
            &{0.2182}&{25.1368}
            &{0.1397}&{27.1394}
            \\ %\hline

            \multicolumn{1}{c}{}&{vox2vec \cite{goncharov2023vox2vec}}
            &{0.2098}&{45.7324}
            &{0.3959}&{38.0991}
            &{0.3609}&{21.0440}
            &\textbf{0.2304}&{25.9604}
            &{0.2247}&\textbf{23.9486}
            &\textbf{0.1443}&{25.5126}
            \\ \cline{2-14}
            
            \multicolumn{1}{c}{}&{Ours} 
            &\textbf{0.2368}&\textbf{44.2361}   
            &\textbf{0.4032}&\textbf{35.3453}       
            &\textbf{0.3745}&\textbf{20.3453}    
            &{0.2253} &\textbf{23.2375} 
            &\textbf{0.2234} &{25.1484} 
            &{0.1402} &\textbf{23.5632} 
            \\ \cline{1-14}
            
            \multicolumn{2}{c}{Sup (50\%)}
            &{0.1882}&{46.3327} 
            &{0.3713}&{36.0685}
            &{0.3557}&{23.8802} 
            &{0.2030}&{23.6540}
            &{0.2354}&{24.7391} 
            &{0.1358}&{22.6528}
            \\ 

            \multicolumn{2}{c}{Sup (100\%)}
            &{0.2398}&{50.0361} 
            &{0.3928}&{34.7032}
            &{0.3962}&{18.8440} 
            &{0.2238}&{19.7038}
            &{0.2477}&{22.0652}  
            &{0.1582}&{20.8641} 
            \\  \hlineB{3} 	
            
        \end{tabular}
    \end{center}
    \vspace{-2em}
\end{table*}
    % \end{small}

\noindent\textbf{Baselines}:
We consider two primary categories of baselines: fully supervised and self-supervised learning.
For fully supervised learning, we use a randomly initialized U-Net architecture without self-supervised pretraining. We train this model from scratch with annotations at 5\%, 10\%, 50\%, and 100\% granularity levels to comprehensively assess the impact of self-supervised pretraining.
As for self-supervised learning, we have two subcategories: contrastive learning and masked image modeling. In contrastive learning, we use SimCLR \cite{chen2020simple}, BYOL \cite{grill2020bootstrap}, SwAV \cite{caron2020unsupervised}, G-L \cite{chaitanya2020contrastive}, {PCRL \cite{zhou2021preservational}, and vox2vec \cite{goncharov2023vox2vec} as baselines for their effectiveness in capturing informative representations. For masked image modeling, we consider state-of-the-art methods such as MAE \cite{he2022masked}, Maskfeat \cite{wei2022masked}, SimMIM \cite{xie2022simmim}, ConvMAE \cite{gao2022convmae}, BEiT V2 \cite{peng2022beit}, BootMAE \cite{dong2022bootstrapped}, and SelfMedMAE \cite{zhou2023self} which are known for modeling complex relationships in medical images through mask-based objectives.
We evaluate the quality of learned representations by transferring them to downstream segmentation tasks, performing end-to-end fine-tuning. This structured approach to baselines ensures a systematic evaluation of AMLP in comparison to a diverse set of benchmarks.

\noindent\textbf{Pre-processing}: 
We apply the following pre-processing steps:
(\romannumeral1) re-sampling of all volumes and corresponding labels to a fixed pixel size $1.0 \times 1.0 \times 1.0$ $mm^{3}$ using nearest-neighbour interpolation, intensity normalization of each 3D volume, clip the value to the $[1, 99]$ percentiles to remove extreme pixel values (such as too high or too low values), avoiding these extreme values from having a negative impact on subsequent processing and model training and normalize the image with mean and std for region nonzero to help the model learn image features more effectively. Therefore, reduce outliers in pixel values and enhance the contrast of the image, making model training more efficient.

(\romannumeral2) all 2D images and corresponding 2D labels are obtained from the z-axis of 3D volumes. The dimension $d$ for each dataset are:
(a) Hecktor: $d$ = $144 \times 144$, and 
(b) BraTS2018: $d$ = $160 \times 160$.

\noindent\textbf{Implementation details}:
Our AMLP is implemented based on Torch 1.6.0 and CUDA-10.1. All experiments are done on $8$ GeForce RTX 2080 GPUs. For self-supervised learning, the AdamW \cite{loshchilov2017decoupled} optimizer is used, with a learning rate of $1e-4$. The temperature parameter $t$ is set to $0.1$. The batch size is $48$ for Hecktor and $36$ for BraTS2018. And a total of $800$ epochs is trained. The patch size of the Hecktor dataset and BraTS2018 dataset images is $16 \times 16$, and the corresponding numbers of the image patches are 81 and 100 respectively. Besides, single-modality data is used for BraTS (T1CE, T2, Flair, and T1) and Hecktor (CT, and PET) datasets.
For transfer learning, the U-Net \cite{ronneberger2015u} for downstream segmentation task is also trained with AdamW optimizer. The initial learning rate is $1e-4$, the weight decay is $0.0001$, and the learning rate strategy is cosine decay. When using $5\%$ labels, batch size is set to $31$ for Hecktor and $56$ for BraTS2018. For $10\%$ labels, batch size is set to $90$ for Hecktor and $70$ for BraTS2018. To make network convergence, $70$ epochs are trained for Hecktor and BraTS2018. 

During implementation, the AMR masking rate starts at 25\% and is gradually increased to 80\% during training. For datasets with extremely subtle lesions, performance can be improved by reducing the initial masking rate and increasing $\tau$. Meanwhile, the MPS strategy assumes that smaller clusters correspond to lesion classes. In atypical cases where lesions occupy most of the image, larger clusters can be assigned to lesion classes.

Besides, all baselines are implemented and run using similar procedures and settings as those in their original papers, where all self-supervised methods are learned in a two-stage learning way, i.e., first pre-trained using only unlabeled data and then fine-tuned using limited labeled data.

\noindent \textbf{Network Architecture}:
We use the U-Net, which includes four downsampling and upsampling modules, as a segmentation backbone network for all methods. Each downsampling module consists of two $3 \times 3$ convolutions and a $2 \times 2$ maxpooling with stride 2, while each upsampling module consists of two $3 \times 3$ convolutions and a $2 \times 2$ transposed convolution with stride 2. This U-Net network also serves as the backbone network of AMLP.

\begin{figure*}[htbp]
    \centering
    % \vspace{-1em}
    \includegraphics[width=0.95\textwidth]{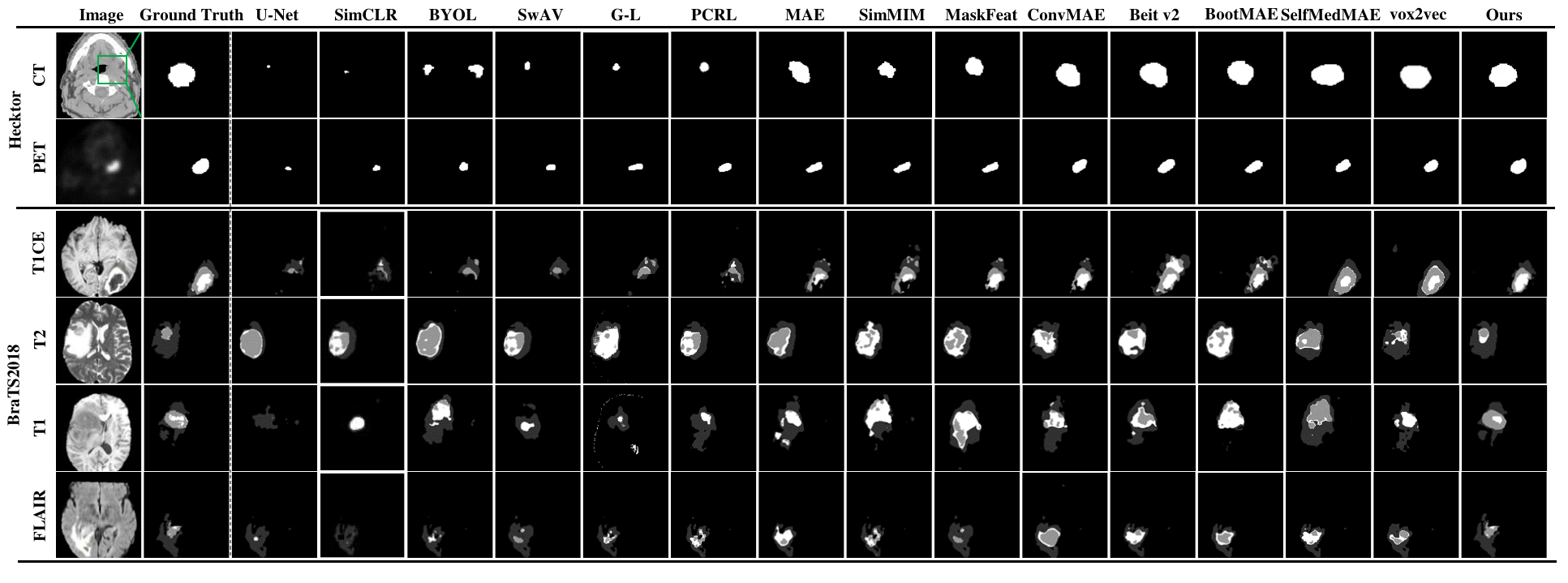}
    \vspace{-0.5em}
    \caption{Visualized segmentation results on the Hecktor and BraTS2018 datasets with $10\%$ labeled data. The first two rows are the results of the Hecktor dataset, specifically the slice number are CHUM044\_slice\_098 (PET) and CHUM012\_slice\_089 (CT), where the head and neck tumor (only one type of tumor) is marked with white in the segmentation results. The last four rows are the results of the BraTS2018 dataset, specifically the slice number are Brats18\_TCIA02\_608\_1\_059 (T1CE), Brats18\_CBICA\_AQO\_1\_084 (T2), Brats18\_CBICA\_AUN\_1\_081 (T1), and Brats18\_2013\_18\_1\_067 (FLAIR), where WT (whole tumor), ET (enhancing tumor), and TC (tumor core) are marked with light gray, medium gray and white, respectively. As observed, our segmentation results are closer to ground truth and better than other methods.}
    \vspace{-1.5em}
    \label{fig:visualization}
\end{figure*}

\noindent \textbf{Fine-tuning}:

In the upstream pre-training task, our model reconstructs masked images. During this stage, the reconstructed images serve only to optimize the network, compelling it to learn more discriminative lesion information and develop robust feature representations for downstream tasks. Subsequently, we transfer the weights of the pre-trained encoder to the downstream model, while its decoder is randomly initialized. For the downstream segmentation task, the model does not use any generated images as input; instead, it performs lesion segmentation directly on real medical images during both the fine-tuning and inference phases.

\noindent\textbf{Evaluation metrics}:
We use four key metrics to evaluate our methods: Dice Similarity Coefficient~(DSC)~\cite{dice1945measures}, Sensitivity (Sen), Mean Intersection of Union Boundaries (BIoU)~\cite{cheng2021boundary}, and 95\% Hausdorff Distance (HD95). DSC and Sen reflect similarity metrics, while BIoU and HD95 measure distance-related aspects. 

Specific formulas for these metrics are as follows: 
\begin{equation}
   DSC = \frac{2 \times TP + \epsilon}{T + P + \epsilon},
\end{equation}

\begin{equation}
   SEN = \frac{TP + \epsilon}{TP + FN + \epsilon},
\end{equation}

\begin{equation}
   BIoU = \frac{(G_d \cap G) \cap (P_d \cap P)}{(G_d \cap G) \cup (P_d \cap P)},
\end{equation}

\begin{equation}
   HD95 = max_{k95\%}[d(P,G), d(G, P)],
\end{equation}

\noindent where $TP$, $FP$, and $FN$ are true positive, false positive, and false negative, respectively. $P$ is the prediction binary mask, $G$ is the ground truth binary mask, $P_d$ and $G_d$ are the sets of pixels in the boundary region of the binary mask.

\vspace{-0.5em}
\subsection{Main Results}

To investigate the effectiveness of our proposed AMLP, we conduct experiments on two datasets and compare its performance with two kinds of state-of-the-art baselines: 
Fully Supervised Baseline (i.e., Fully Supervised) and Self-Supervised Baselines (i.e., SimCLR, BYOL, SwAV, MAE, Maskfeat, SimMIM, ConvMAE, BEiT v2, BootMAE). 
For a fair comparison, we use the same backbone network (U-Net) with 5\% and 10\% annotations across all methods. The experimental results are shown in Table \ref{table:main_1} and Table \ref{table:main_2}, which respectively refer to two types of indicators: similarity metrics (i.e., DSC and Sen) and distance metrics (i.e., BIoU and HD95). 
For ease of comparison, we report only the average results for different modalities of the BraTS2018 dataset in the main experimental results. This average is obtained by calculating the individual segmentation results for the three different hierarchical labels and then averaging them.
The visualization of the segmentation results of AMLP and other methods are shown in Fig. \ref{fig:visualization}.

As shown in Table \ref{table:main_1} and Table \ref{table:main_2}, AMLP generally outperforms all the state-of-the-art self-supervised medical image segmentation baselines in terms of all evaluation metrics and two small label-ratio settings (i.e., 5\% and 10\%) on both datasets, and its performances are similar to (and sometimes even better than) the fully supervised solution using much higher ratios (i.e., 50\% and 100\%) of label data. This observation proves that the proposed AMLP can achieve superior medical image segmentation performances using only a small number of annotations, which thus greatly reduces the labeling workload of applying intelligent medical image segmentation systems in clinical practices. Detailed analysis is as follows.

\noindent\textbf{Comparison with state-of-the-art fully-supervised methods.}
As shown in Table \ref{table:main_1} and Table \ref{table:main_2}, self-supervised learning methods (including the proposed AMLP) achieve much better segmentation performances than the fully supervised baseline with the same ratio of labeled data. This enhancement stems from the ability of self-supervised methods to extract valuable insights from the unlabeled data.

Furthermore, AMLP generally outperforms the baseline model that is fully supervised learning from scratch by a large margin with $5\%$ and $10\%$ annotations. The performance of AMLP with $5\%$ annotations can be close to or even better than that of the  fully supervised method with $50\%$ annotations. As for $10\%$ annotations, AMLP not only outperforms the fully supervised method with $50\%$ annotations but also outperforms the fully supervised method with $100\%$ annotations in the PET modality on Hecktor. Specifically, in the case of 5\% annotation on the CT modality of the Hecktor dataset, we observe that AMLP outperforms the fully supervised method by 11.65\% for DSC, 10.1\% for Sen, and 8.24\% for BIoU, and 4.4292 for HD95; while on the T1 modality of the BraTS2018 dataset, AMLP outperforms the fully supervised method by 15.54\% for DSC, 23.21\% for Sen, and 6.73\% for BIoU, and 9.3621 for HD95. This is because the ability of our method to learn more valuable representations from a large amount of unlabeled data, thereby improving the performance of downstream segmentation models.

\begin{table*}[!t] 
    \begin{center}
        % \vspace{-0.5em}
        \caption{Results of our ablation studies on the Hecktor and BraTS2018 datasets with 5\% and 10\% labelled data.}
        \vspace{-0.5em}
        \smallskip 
        \label{table:ablation}
        \setlength\tabcolsep{2pt}
        \newcolumntype{"}{@{\hskip\tabcolsep\vrule width 1pt\hskip\tabcolsep}}
        \begin{tabular}{m{0.05\textwidth}<{\centering}m{0.10\textwidth}<{\centering}m{0.06\textwidth}<{\centering}p{0.06\textwidth}<{\centering}p{0.06\textwidth}<{\centering}p{0.06\textwidth}<{\centering}p{0.06\textwidth}<{\centering}p{0.06\textwidth}<{\centering}p{0.06\textwidth}<{\centering}p{0.06\textwidth}<{\centering}p{0.06\textwidth}<{\centering}p{0.06\textwidth}<{\centering}p{0.06\textwidth}<{\centering}p{0.06\textwidth}<{\centering}}
            \hlineB{3}
            \multicolumn{2}{c}{\textbf{Methods}}
            & \multicolumn{4}{c}{\textbf{Hecktor}}     
            & \multicolumn{8}{c}{\textbf{BraTS2018}}  \\    
            % \multicolumn{2}{c}{}&\multicolumn{4}{c}{}&\multicolumn{8}{c}{}\\ 
            \cmidrule(lr){3-6} \cmidrule(lr){7-14}
            
            \multicolumn{2}{c}{Modality} &\multicolumn{2}{c}{CT} & \multicolumn{2}{c}{PET}& \multicolumn{2}{c}{T1CE}& \multicolumn{2}{c}{T2} 
            & \multicolumn{2}{c}{FLAIR} & \multicolumn{2}{c}{T1}   \\ \cmidrule(lr){3-4} \cmidrule(lr){5-6} \cmidrule(lr){7-8} \cmidrule(lr){9-10} \cmidrule(lr){11-12} \cmidrule(lr){13-14}
            
            \multicolumn{2}{c}{Metrics} 
            &{DSC$\uparrow$}&{HD95$\downarrow$}  &DSC$\uparrow$&{HD95$\downarrow$}  &DSC$\uparrow$&{HD95$\downarrow$}  &DSC$\uparrow$&{HD95$\downarrow$}   &DSC$\uparrow$&{HD95$\downarrow$}  &DSC$\uparrow$&\multicolumn{1}{c}{HD95$\downarrow$}\\ \cline{1-14}

            \specialrule{0.5pt}{0pt}{0pt}
              
            \multirow{9}{*}{\textbf{5\%}}{} &{Sup} 
            & {0.1740} & {62.7115} & {0.5452} & {69.0963} & {0.3234} & {55.0727} & {0.3952} & {47.4584} & {0.4035} & {33.2086} & {0.1743}  & {59.5963}\\ 
    
            \multicolumn{1}{c}{}&{BASE} &{0.2373} & {61.2351} & {0.5692} & {62.3420} & {0.5307} & {53.7239} & {0.4152} & {45.2062} & {0.4401} & {32.6258} & {0.2486} & {57.7398}\\ 

            \multicolumn{1}{c}{}&{MPS} &{0.2679} & {60.0462} & {0.5921} & {58.6432} & {0.5501} & {51.3462} & {0.4396} & {43.6237} & {0.4581} & {30.9462} & {0.2791} & {55.0862}\\ 

            \multicolumn{1}{c}{}&{AMR} &{0.2512} & {61.2337} & {0.5834} & {60.7285} & {0.5425} & {52.6132} & {0.4285} & {44.9625} & {0.4423} & {31.4195} & {0.2673} & {56.6478}\\ 
            
            \multicolumn{1}{c}{}&{CRCL} &{0.2481} & {60.7930} & {0.5802} & {61.1427} & {0.5399} & {52.0410} & {0.4205} & {44.8291} & {0.4471} & {30.4632} & {0.2578} & {56.9203}\\ 
            
            \multicolumn{1}{c}{}&{MPS-AMR} &{0.2831} & {58.6127} & {0.6244} & {43.4581} & {0.5752} & {45.4307} & {0.4693} & \textbf{40.4502} & {0.4687} & {28.4192} & {0.3143} & {50.9173}\\  
            
            \multicolumn{1}{c}{}&{MPS-CRCL} &{0.2804} & {59.3581} & {0.6114} & {51.2614} & {0.5637} & {48.7651} & {0.4563} & {41.5846} & {0.4649} & {29.6834} & {0.3062} & {52.3516}\\ 
            
            \multicolumn{1}{c}{}&{AMR-CRCL} &{0.2816} & {60.0574} & {0.6044} & {55.8031} & {0.5580} & {50.0917} & {0.4409} & {42.9724} & {0.4600} & {30.1734} & {0.2916} & {53.4329}\\ 

            \multicolumn{1}{c}{} &{AMLP(Ours)} &\textbf{0.2905} & \textbf{57.2823} & \textbf{0.6296} & \textbf{40.5376} &\textbf {0.5762} & \textbf{44.2326} & \textbf{0.4721} & {40.6362}& \textbf{0.4693} & \textbf{28.2352} & \textbf{0.3297} & \textbf{50.2342}\\

            \specialrule{0.5pt}{0pt}{0pt}
            
            \multirow{9}{*}{\textbf{10\%}}{} &{Sup} & {0.2541} & {64.0604} & {0.5769} & {41.8627} & {0.4451} & {50.9107} & {0.4288} & {43.9483} & {0.4489} & {30.4918} & {0.2476}  & {52.8203}\\ 
    
            \multicolumn{1}{c}{}&{BASE} &{0.3006} & {58.2945} & {0.5932} & {49.6235} & {0.5892} & {47.6230} & {0.4518} & {40.4253} & {0.4599} & {29.0526} & {0.3351} & {45.2526}\\ 

            \multicolumn{1}{c}{}&{MPS} &{0.3318} & {55.4185} & {0.6079} & {47.2145} & {0.6067} & {32.4192} & {0.4827} & {31.2364} & {0.4823} & {27.2345} & {0.3576} & {35.4273}\\ 

            \multicolumn{1}{c}{}&{AMR} &{0.3187} & {57.6234} & {0.5958} & {48.5263} & {0.5903} & {41.5234} & {0.4735} & {38.6234} & {0.4748} & {28.9632} & {0.3469} & {39.9634}\\ 
            
            \multicolumn{1}{c}{}&{CRCL} &{0.3105} & {57.7369} & {0.5960} & {49.1352} & {0.5883} & {45.8265} & {0.4693} & {39.6515} & {0.4701} & {28.5238} & {0.3390} & {41.3282}\\ 
            
            \multicolumn{1}{c}{}&{MPS-AMR} &{0.3671} & {46.5192} & {0.6303} & {37.4192} & {0.6332} & {22.3254} & {0.5157} & {24.2163} & {0.4967} & {25.4163} & {0.3807} & {24.2154}\\ 
            
            \multicolumn{1}{c}{}&{MPS-CRCL} &{0.3532} & {50.2356} & {0.6221} & {39.7684} & {0.6256} & {26.5163} & {0.5012} & {27.3175} & {0.4912} & {26.4192} & {0.3732} & {29.6234}\\ 
            
            \multicolumn{1}{c}{}&{AMR-CRCL} & {0.3481} & {51.3252} & {0.6185} & {42.4560} & {0.6185} & {30.5124} & {0.4976} & {30.2864}& {0.4891} & {27.4258} & {0.3701} & {30.5128}\\ 

            \multicolumn{1}{c}{} &{AMLP(Ours)} &\textbf{0.3690} & \textbf{44.2361} & \textbf{0.6314} & \textbf{35.3453} & \textbf{0.6343} & \textbf{20.3453} & \textbf{0.5194} & \textbf{23.2375}& \textbf{0.4971} & \textbf{25.1484} & \textbf{0.3815} & \textbf{23.5632}\\ 

            \specialrule{1.0pt}{-0.09em}{-0.2em}
            
        \end{tabular}
    \end{center}
    \vspace{-1em}
\end{table*}

\begin{table*}[!t]
\begin{adjustwidth}{-0.35in}{-0.35in}
\begin{center}
\caption{Results of loss-based ablation studies on the Hecktor and BraTS2018 datasets with $5\%$ labeled data.}
\vspace{-0.5em}
\smallskip 
\label{table:loss_effective}
\resizebox{0.9\linewidth}{!}{
  \begin{tabular}{*{13}{c}}
  \specialrule{1.0pt}{0pt}{0.05ex}
  \multirow{3}*{\textbf{Losses}} & 
  \multicolumn{4}{c}{\textbf{Hecktor}} & \multicolumn{8}{c}{\textbf{BraTS2018}} \\ 
  \multicolumn{1}{c}{{{}}} & \multicolumn{2}{c}{{{\textbf{CT}}}} & \multicolumn{2}{c}{{{\textbf{PET}}}} & \multicolumn{2}{c}{{{\textbf{T1CE}}}} & \multicolumn{2}{c}{{{\textbf{T2}}}}& \multicolumn{2}{c}{{{\textbf{FLAIR}}}}& \multicolumn{2}{c}{{{\textbf{T1}}}} \\ \cline{2-13}
  
  & \multicolumn{1}{c}{DSC$\uparrow$} & \multicolumn{1}{c}{HD95$\downarrow$} & \multicolumn{1}{c}{DSC$\uparrow$} & \multicolumn{1}{c}{HD95$\downarrow$} & \multicolumn{1}{c}{DSC$\uparrow$} & \multicolumn{1}{c}{HD95$\downarrow$} & \multicolumn{1}{c}{DSC$\uparrow$} & \multicolumn{1}{c}{HD95$\downarrow$} & \multicolumn{1}{c}{DSC$\uparrow$} & \multicolumn{1}{c}{HD95$\downarrow$} & \multicolumn{1}{c}{DSC$\uparrow$} & \multicolumn{1}{c}{HD95$\downarrow$}\\

  \specialrule{0.5pt}{0pt}{0pt}
  \multirow{1}{*}
  {$L_{pred}$} &{0.2373} & {61.2351} & {0.5692} & {62.3420} & {0.5307} & {53.7239} & {0.4152} & {45.2062}& {0.4401} & {32.6258} & {0.2486} & {57.7398}\\  \cline{1-13}
  
  {$L_{pred}$}$+L_{rrl}$ & {0.2863} & {59.2341} & {0.6152} & {49.2341} & {0.5752} & {47.1562} & {0.4673} & {42.1345}& {0.4617} & {29.9632} & {0.3187} & {53.2635}\\ 
  
  {$L_{pred}$}$+L_{ccl}$ & {0.2592} & {60.4562} & {0.5745} & {55.4562} & {0.5343} & {53.5263} & {0.4231} & {44.5263}& {0.4462} & {32.3526} & {0.2614} & {56.8561}\\
  
  {$L_{pred}$}$+L_{rrl}+L_{ccl}$ (Ours)&{\textbf{0.2905}} & {\textbf{58.2823}} & {\textbf{0.6296}} & {\textbf{40.5376}} & {\textbf{0.5762}} & {\textbf{44.2326}}  & {\textbf{0.4721}} & {\textbf{40.2362}}& {\textbf{0.4693}}  & {\textbf{28.2352}} & {\textbf{0.3297}}  & {\textbf{50.2342}}\\
  
  \specialrule{1.0pt}{-0.09em}{-0.2em}
 \end{tabular}
 }
 \vspace{-1em}
\end{center}
\end{adjustwidth}
\end{table*}

\noindent\textbf{Comparison with state-of-the-art self-supervised methods.}
In Table \ref{table:main_1} and Table \ref{table:main_2}, we compared AMLP with state-of-the-art self-supervised methods. AMLP consistently outperforms these methods across nearly all cases on both datasets, which demonstrates AMLP's ability to extract more comprehensive information and features from unlabeled multimodal data compared to current self-supervised contrastive learning baselines, resulting in superior medical image segmentation. For example, in the case of 5\% annotation on CT modality of the Hecktor dataset, we observe that AMLP outperforms the fully supervised method by 1.46\% for DSC, 1.94\% for Sen, and 1.4\% for BIoU; while on T1 modality of the BraTS2018 dataset, AMLP outperforms the fully supervised method by 2.7\% for DSC, 1.32\% for Sen, and 0.45\% for BIoU, and 0.0289 for HD95. The superior performance of AMLP can be attributed to the fact that it determines more appropriate and challenging adjustable masks for images. Reducing the uncertainty of masked patches and increasing the upper limit of conditional mutual information enables the acquisition of more comprehensive and effective representation information.

While AMLP generally excels, we also observe that its results may not always be the best in few scenarios. This can attribute to the challenge of fully capturing all informative representations with limited labeled data during self-supervised pretraining. Future research could explore the integration of attention mechanisms, such as those in \cite{Shijie-TNSE2024, XU2023107149}, to enhance model performance by focusing on the foreground area, potentially closing any remaining performance gaps even with limited annotations.

\noindent\textbf{Analysis of visualized segmentation results.}
Moreover, these findings align with the visualized results in Fig. \ref{fig:visualization}, where AMLP are closer to ground truth and better than other self-supervised medical image segmentation methods. A comprehensive analysis is as follows:

The contrastive methods like SimCLR, BYOL, and SwAV often lead to over-segmentation and imprecise lesion contour delineation, indicating their difficulty in preserving sharp boundaries and accurate localization due to a focus on global features.
Then, the reconstruction-based approaches such as MAE, MaskFeat, and SimMIM offer finer segmentation details but still exhibit margin blurring and occasional leakage into adjacent tissues, suggesting that reconstruction alone is insufficient for precise lesion delineation.
In contrast, AMLP produces segmentations closest to the ground truth, with sharper contours,  and improved segmentation of small lesions, such as the enhancing ring in the T1CE slice.

This visual evidence strongly underscores AMLP's advantages, which stem from its ability to focus selectively on lesion patches and optimize representation learning, which enables the model to learn more discriminative and robust features for accurate medical image segmentation.

\vspace{-1em}
\subsection{Ablation Studies}
We conducted ablation studies to evaluate the effectiveness of the four key enhancements: the masked patch selection strategy, relative reconstruction loss, category consistency loss, and adjustable masking ratio. Specifically, we have implemented several models for self-supervised pretraining: 
(\romannumeral1) BASE: The BASE model is a self-supervised learning model, encompassing the pre-training upstream model (masking and reconstruction task), weight transferring, and fine-tuning of the downstream model (segmentation task). These processes are implemented based on the U-Net network.
% BASE is a model that combines the random masking strategy with the U-Net as the reconstruction model for self-supervised pretraining.}
(\romannumeral2) MPS: MPS incorporates only the masked patches selection strategy on the basis of BASE. 
(\romannumeral3) AMR: AMR incorporates only the adjustable masking ratio strategy on the basis of BASE. 
(\romannumeral4) CRCL: Combining the attention reconstruction loss and category consistency loss on the basis of BASE.
(\romannumeral5) MPS-AMR: Integrating both the masked patches selection strategy and adjustable masking ratio strategy on the basis of BASE.
(\romannumeral6) MPS-CRCL: Combining the masked patches selection strategy with the joint losses (CRCL) on the basis of BASE.
(\romannumeral7) AMR-CRCL: Incorporating the adjustable masking ratio strategy along with the combined losses (CRCL) on the basis of BASE.
(\romannumeral8) AMLP (Ours): Represent the full model, incorporating all improvements on the basis of BASE.

\begin{table*}[!t]
\begin{adjustwidth}{-0.4in}{-0.4in}
\begin{center}
% \vspace{1em}
\caption{Results of different masking strategies on the Hecktor and BraTS2018 datasets with $10\%$ labeled data.}
\vspace{-0.5em}
\smallskip 
\label{table:different masking strategies}
\renewcommand\arraystretch{1.2}
\resizebox{0.9\linewidth}{!}{
  \begin{tabular}{*{13}{c}}
  \specialrule{1.0pt}{0pt}{0.05ex}
  \multirow{3}*{\textbf{Losses}} & 
  \multicolumn{4}{c}{\textbf{Hecktor}} & \multicolumn{8}{c}{\textbf{BraTS2018}} \\ 
  \multicolumn{1}{c}{{{}}} & \multicolumn{2}{c}{{{\textbf{CT}}}} & \multicolumn{2}{c}{{{\textbf{PET}}}} & \multicolumn{2}{c}{{{\textbf{T1CE}}}} & \multicolumn{2}{c}{{{\textbf{T2}}}}& \multicolumn{2}{c}{{{\textbf{FLAIR}}}}& \multicolumn{2}{c}{{{\textbf{T1}}}} \\ \cline{2-13}
  & \multicolumn{1}{c}{DSC$\uparrow$} & \multicolumn{1}{c}{HD95$\downarrow$} & \multicolumn{1}{c}{DSC$\uparrow$} & \multicolumn{1}{c}{HD95$\downarrow$} & \multicolumn{1}{c}{DSC$\uparrow$} & \multicolumn{1}{c}{HD95$\downarrow$} & \multicolumn{1}{c}{DSC$\uparrow$} & \multicolumn{1}{c}{HD95$\downarrow$} & \multicolumn{1}{c}{DSC$\uparrow$} & \multicolumn{1}{c}{HD95$\downarrow$} & \multicolumn{1}{c}{DSC$\uparrow$} & \multicolumn{1}{c}{HD95$\downarrow$}\\

  \specialrule{0.5pt}{0pt}{0pt}
  \multirow{1}{*}
  {random} &  {0.3256} & {50.2351} & {0.6156}  & {45.2365} & {0.6038} & {30.2532} & {0.4957} & {30.3209} & {0.4624} & {28.2390} & {0.3425} & {35.2624} \\ 
  
  {easy to hard} &  {0.3487} & {47.2630} & {0.6202} & {40.9863} & {0.6195} & {25.2263} & {0.5072} & {27.0117} & {0.4782} & {26.8805} & {0.3752} & {28.8764} \\ 
  
  {hard to easy (Ours)}&\textbf{0.3690} & \textbf{44.2361} & \textbf{0.6314} & \textbf{35.3453} & \textbf{0.6343} & \textbf{20.3453} & \textbf{0.5194} & \textbf{23.2375} & \textbf{0.4971} & \textbf{25.1484} & \textbf{0.3815} & \textbf{23.5632} \\
  \specialrule{1.0pt}{-0.09em}{-0.2em}
 \end{tabular}
 }
\vspace{-1.5em}
\end{center}
\end{adjustwidth}
\end{table*}

The ablation studies are conducted on Hecktor and BraTS2018 datasets using 5\% and 10\% ratios of annotations, and the corresponding results are shown in Table \ref{table:ablation}. Additionally, loss ablation experiments are conducted on the two datasets with 5\% labeled data to assess the effectiveness of different losses, and the results are detailed in Table \ref{table:loss_effective}.

\noindent\textbf{Effectiveness of masked patches selection.}
First, to verify whether k-means can effectively distinguish between foreground and background, we studied its classification precision at the beginning of the training. We randomly selected 150 samples from the BraTS2018 dataset and statistically analyzed the k-means classification results (k=2) for each sample, recording their averages in Table \ref{tab:Precision}. Specifically, each sample was divided into 100 image patches of size $16 \times 16$. According to the statistical results, category 1 contains an average of 15.47 image patches, with an average of 7.39 patches being lesion patches, resulting in a classification precision of 47.77\% for lesion patches in this category. Category 2 contained an average of 84.53 image patches, with an average of 4.88 patches being lesion patches, resulting in a classification precision of 5.77\%. We find that the classification precision of lesion patches in category 1 is much higher than in category 2, indicating that category 1 contains more lesion patches, while category 2 mainly contains background patches. Therefore, we can use k-means as the basic method for distinguishing between foreground and background and further optimize this method during training to improve classification precision. Additionally, we observed that the number of patches in category 1 is always much smaller than in category 2. Since lesion areas in medical images occupy only a small part of the entire image, our method defaults to considering the category with fewer patches as the foreground and the category with more patches as the background.

\begin{table}[!t]
    \caption{Patches' statistical results from k-means clustering. We report the average number of patches and lesion patches over 150 samples. We found that category 1 is much better at selecting lesion patches than category 2, which verifies our motivation and model design.}
    \centering
    \resizebox{0.47\textwidth}{!}{
        \begin{tabular}{cccc}
            \toprule
            \multicolumn{1}{c}{} & \multicolumn{1}{c}{Number of patches} & \multicolumn{1}{c}{Number of lesion patches} & Precision  \\
            \midrule
            category 1 & 15.47 & 7.39 & 47.77\% \\
            category 2 & 84.53 & 4.88 & 5.77\% \\
            \bottomrule
        \end{tabular}}
    \centering
    \label{tab:Precision}
    \vspace{-2.3em}
\end{table}

Based on this, we further verified the effectiveness of the MPS strategy, with results shown in Table \ref{table:ablation}. The results demonstrate that MPS consistently outperforms BASE. For instance, employing only MPS in Hecktor CT with 5\% labeled data increased DSC by 3.06\% and reduced HD95 by 1.1889 compared to BASE. removing MPS from the complete AMR-CRCL model resulted in performance degradation with DSC decreasing by 0.89\% and HD95 increasing by 2.7751. This can be attributed to MPS's targeted focus on lesion patches during reconstruction, enabling better capture of lesion texture and shape characteristics. This targeted learning approach proves particularly beneficial for medical images where lesions often present complex shapes and blurred boundaries.

\noindent\textbf{Effectiveness of CRCL}
To evaluate the effectiveness of the combined losses, we conducted loss-based ablation studies on Hecktor and BraTS2018 datasets with 5\% and 10\% labeled data. The combined losses refer to the relatvie reconstruction loss $L_{rrl}$ and category consistency loss $L_{ccl}$. Since $L_{ccl}$ depends on the MPS, so it is excluded when evaluating the part that does not contain MPS. 

As shown in Table \ref{table:loss_effective}, the results show that using $L_{rrl}$ or $L_{ccl}$ alone can improve model performance. In the Hecktor CT modality with 5\% labeled data, using $L_{rrl}$ alone can improve DSC by 4.9\%, and using $L_{ccl}$ alone can reduce HD96 by 0.7789. This is because $L_{rrl}$ can help model focus on lesion area accurately, and $L_{ccl}$ can alleviate the misclassification problem of unsupervised k-means and enhance the model's ability to distinguish between foreground and background. At the same time, we also found that the combination of $L_{rrl}$ and $L_{ccl}$ can further improve model performance.

\noindent\textbf{Effectiveness of adjustable masking ratio.}
Besides, as shown in Table \ref{table:ablation}, we can observe that the results of AMR is generally better than BASE. For instance, using AMR on BraTS2018 T1 with 5\% labeled data increased DSC by 1.87\% and reduced HD95 by 1.092. And by removing AMR from the whole model, which is CCL(-MPS-RRL), we can also see the results of DSC reduce by 2.35\% and HD95 for 2.1174. This boost results from AMR's adjustable masking schedule, promoting continuous representation learning, facilitating the acquisition of richer and more diverse representations.

\noindent\textbf{Effectiveness of masking strategies.}
Furthermore, to verify the effectiveness of masking strategy, we design additional ablation study in Table \ref{table:different masking strategies}.
Specifically, we compare three kinds of masking strategies: 
(\romannumeral1) random masking: randomly mask image patches.
(\romannumeral2) easy to hard: preferentially mask background patches, then gradually increase masking ratio.
(\romannumeral3) hard to easy: preferentially mask lesion patches, then gradually increase masking ratio.

The results shows that the performance of ``hard to easy" is superior to ``easy to hard", with ``random" yielding the poorest result, indicating that masking lesion areas firstly leads to better performance than masking background. For example, in the Hecktor CT mode with 10\% labeled data, the DSC result of ``hard to easy" is 2.03\% higher than that of ``easy to hard" and 4.34\% higher than ``random".
This improvement can be attributed to the fact that lesion areas contain critical information necessary for accurate segmentation. By focusing on masking these regions first, the model is forced to learn more robust and discriminative features to handle the most challenging parts of the image. Simultaneously, this strategy is also consistent with the typical diagnostic process, where radiologists typically first examine the lesion and then consider the condition of the surrounding tissues. On the other hand, starting with background masking might lead to less informative learning initially, as the background regions generally contain less complex features compared to lesions. Random masking, lacking any strategic prioritization, results in the worst performance, highlighting the importance of a well-thought-out masking strategy in training effective models.

\begin{table}[!t]
    \caption{The DSC results of different clustering methods achieved by AMLP on T1 modality of BraTS2018 with 10\% labeled data.}
    \centering
    \label{table_clustering}
    \begin{tabular}{ccccc}
    \toprule
    methods & k-means & hierarchical & t-SNE &  DBSCAN \\
    \midrule
    DSC & 0.3815 & 0.3659 & 0.3894 & 0.3770 \\
    Complexity & $O(n)$ & $O(n^2)$ & $O(n^2)$ & $O(n^2)$ \\
    \bottomrule
    \end{tabular}
    \vspace{-1.5em}
\end{table}

\begin{figure}[!t]
	\centering
	\subfigure[Hecktor dataset]{
		\begin{minipage}[b]{0.231\textwidth}
			\includegraphics[width=1\textwidth]{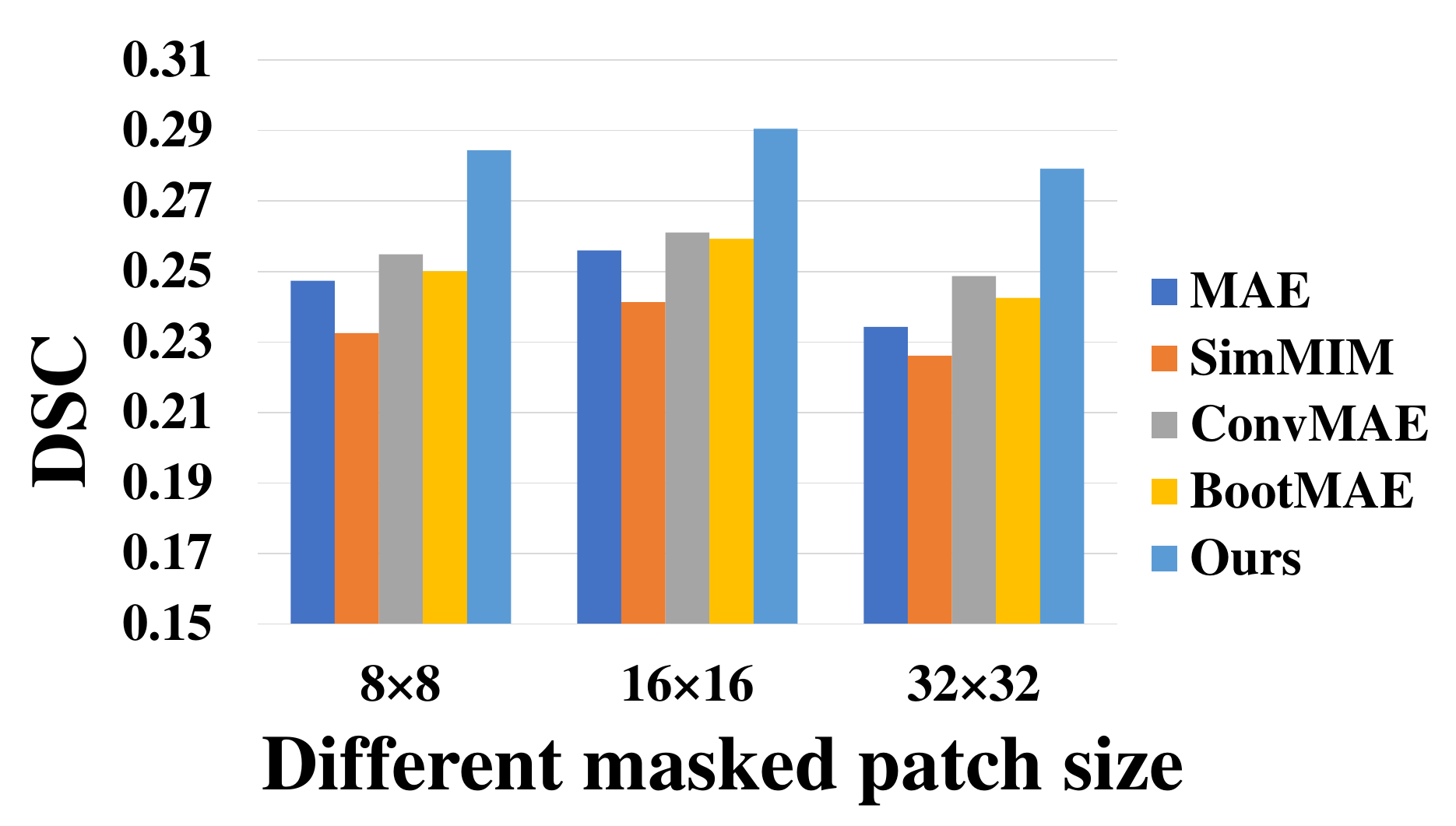}
		\end{minipage}
		\label{fig:size_hecktor}
	}
    	\subfigure[BraTS2018 dataset]{
    		\begin{minipage}[b]{0.223\textwidth}
   		 	\includegraphics[width=1\textwidth]{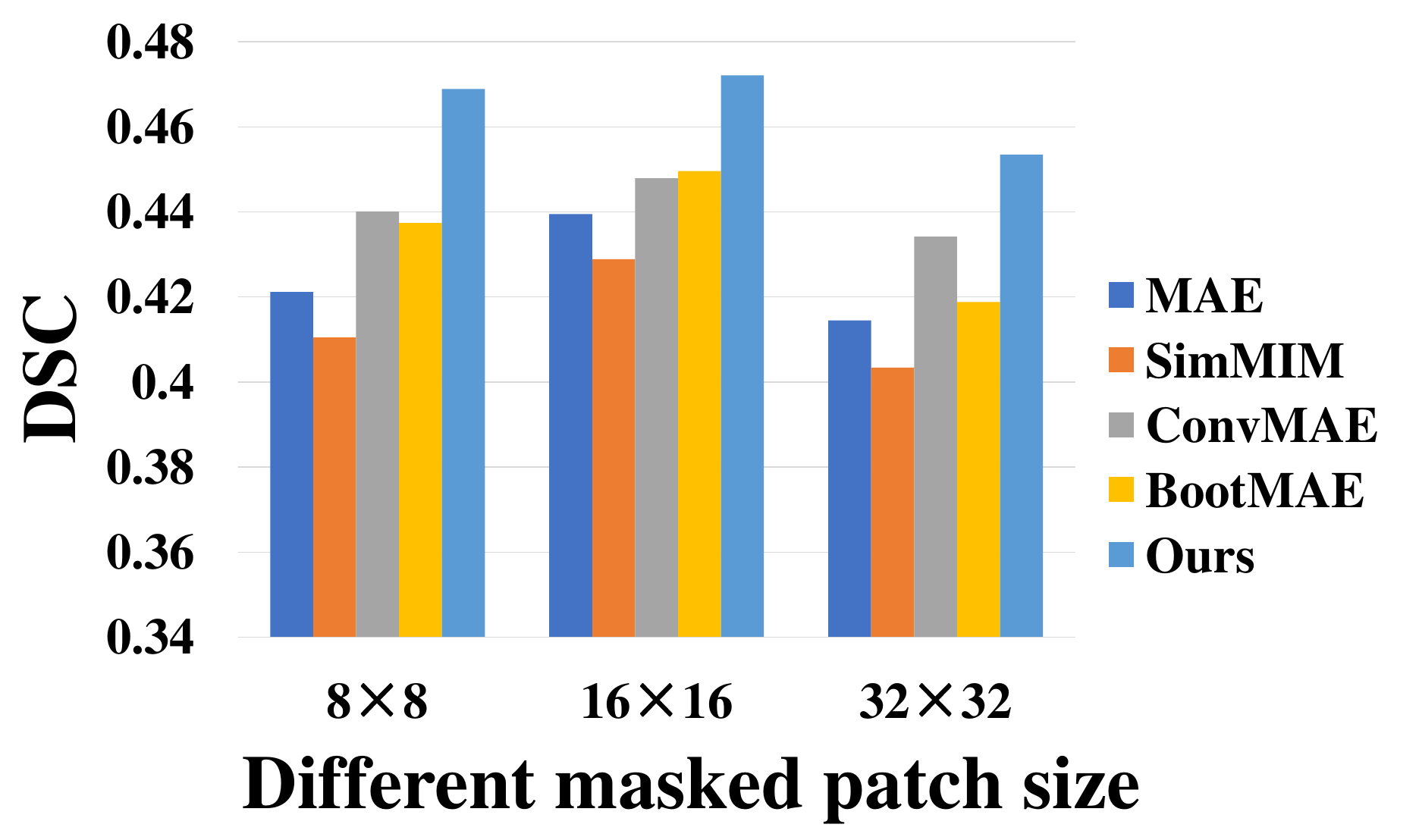}
    		\end{minipage}
		\label{fig:size_BraTS2018}
    	}
    \vspace{-0.5em}
	\caption{Compare the DSC segmentation results of the proposed AMLP and other MIM methods on different masked patch sizes on the CT modality of Hecktor and T2 modality of BraTS2018 datasets with 5\% labeled data.}
	\label{fig:patch_size}
     \vspace{-1.5em}
\end{figure}

\vspace{-0.5em}
\subsection{Additional Experiments}

\noindent\textbf{Analysis of different clustering methods.}
To determine the optimal clustering approach for lesion segmentation, we evaluated four prevalent unsupervised methods on AMLP and T1 modality of BraTS2018 dataset with 10\% labeled data: k-means, hierarchical clustering, t-SNE and DBSCAN. As shown in Table \ref{table_clustering}, the DSC k-means achieved is 0.3815. This result was only 0.79\% lower than the best score of 0.3894 attained by t-SNE. However, k-means held a significant advantage in computational complexity, with a time complexity of $O(n)$ compared to $O(n^2)$ for other methods. The combination of strong performance approaching the optimum score with low complexity highlights k-means as an efficient and effective choice compatible with the hardware and time constraints of real-world clinical deployment. Compared to hierarchical and density-based clustering, k-means provides a favorable trade-off between accuracy and efficiency critical for practical utility.

\noindent\textbf{Analysis of different masked patch sizes.}
Furthermore, in order to explore the impact of patch size on model performance and to find the most suitable patch size for medical image analysis tasks, we conducted additional experiment to compare the DSC performance of AMLP and other baselines using different masked patch size of $8 \times 8$, $16 \times 16$ (used in our paper) and $32 \times 32$ on the CT modality of Hecktor dataset and T2 modality of BraTS2018 dataset with 5\% labeled data.
Results are shown in Fig. \ref{fig:patch_size}.
The results indicates that our proposed AMLP outperforms other baselines across different masked patch sizes on both datasets, demonstrating the effectiveness of our proposed AMLP in medical image segmentation task. Additionally, all methods achieve the best performance with the masked patch of 16x16, suggesting that a moderate patch size is optimal in MIM methods. This is because smaller patch sizes may cause the model to learn too many short connections, and too many corresponding hints from smaller visible patches may limit the improvement in generalization ability. On the other hand, larger masked patch size may enforce the network to learn relatively long-range connections and make the learning process too difficult.

\begin{table}[!t]
    \caption{DSC results with different pre-set misclassification ratios on the T2 modality of BraTS2018 using 10\% labeled data. ``w/ updated patch labels" means the pre-classified incorrect label categories updates with the training process. ``w/o updated patch labels" means unchanged.}
    \centering
    \label{mis_patch}
    \begin{tabularx}{0.47\textwidth}{ccccc}
    \toprule
    pre-set misclassification ratio & 0\% & 10\% & 20\% & 30\%  \\
    \midrule
    w/ updated patch labels & 0.5194 & 0.5150 & 0.5103 & 0.4975 \\
    % \midrule
    w/o updated patch labels & 0.5194 & 0.5059 & 0.4932 & 0.4740 \\
    \bottomrule
    \end{tabularx}
    \vspace{-1.5em}
\end{table}

\noindent\textbf{Analysis of tolerance for misclassification.}
To validate our proposed model's ability to correct initial misclassifications, we have conducted pre-set misclassification ratios experiments as shown in ``w/ updated patch labels" of Table \ref{mis_patch}. Specifically, before training commenced, we deliberately set a fixed proportion (0\%, 10\%, 20\%, and 30\%) of misclassified image patches in the input to the network, allowing the network to update the class labels of all image patches during training. We then tested the model's segmentation performance. The results indicate that the DSC for the cases with 10\% and 20\% pre-misclassified patches only differed by 0.44\% and 0.91\%, respectively, from the 0\% misclassification scenario. This suggests that the model can correct initial misclassifications to some extent. However, when the pre-misclassification rate reached 30\%, the performance difference compared to 0\% misclassification increased to 2.19\%, highlighting the potential impact of misclassification rates on model performance.

Additionally, to investigate our method's tolerance to misclassification, we conducted experiments where the proportion of misclassified patches was also fixed throughout the training process, meaning that the network could not update the class labels of these pre-misclassified patches. The results, shown in  Table \ref{mis_patch} under the ``w/o updated patch labels" condition, reveal that the performance with a 10\% initial misclassification rate differed from the 0\% rate by only 1.35\%. This indicates that even with 10\% of patches misclassified, the performance loss is relatively limited, demonstrating the model's tolerance to misclassification. However, when the predetermined misclassification rates reached 20\% or 30\%, the DSC differences compared to 0\% were 2.62\% and 4.54\%, respectively. This shows that the model's tolerance decreases as the misclassification rate increases.

However, when verifying the effectiveness of k-means, we previously found that even an untrained k-means algorithm has a maximum misclassification rate of approximately 12.96\% for foreground-background classification, as shown in  Table \ref{tab:Precision}. This means that our model is highly unlikely to encounter a misclassification rate as high as 20\% or 30\%, but rather around 10\% or even lower. Therefore, our model has sufficient corrective ability and fault tolerance in most cases, ensuring the stability of our model's performance.

\noindent\textbf{Analysis of single hierarchical labels segmentation.}
We further present all three single-class DSC results of T1CE modality from BraTS2018 dataset in Table \ref{three_tumor-region}. With 5\% labeled data, our method not only outperforms the average results of three levels of labels but also consistently surpasses other methods in all single tumor categories. Specifically, our method achieves 1.39\%, 0.75\%, and 0.68\% better DSC in the segmentation of WT, TC, and ET structures, respectively, compared to the second-best results. This is attributed to our method's stronger ability to distinguish between lesion and background areas, leading to more accurate results in various lesion edge segmentations.

\begin{table}[!t]
    \caption{Comparison of DSC segmentation results for each single hierarchical label in the T1CE modality of the BraTS2018 dataset with 5\% labeled data. Among them, WT represents the whole tumor, TC represents the tumor core, ET represents the enhancing tumor, and Avg. indicates the mean DSC of the segmentation results for the three categories.}
    \centering
    \label{three_tumor-region}
    \begin{tabularx}{0.4\textwidth}{Xcccc}
    \toprule
    Methods & ET & WT & TC &  Avg. \\
    \midrule
    SimCLR & 0.4299 & 0.6540 & 0.5370 & 0.5403 \\
    BYOL & 0.4085 & 0.6391 & 0.5259 & 0.5245 \\
    SwAV & 0.4301 & 0.6215 & 0.5240 & 0.5252 \\
    MAE & 0.4337 & 0.6667 & 0.5391 & 0.5465 \\
    SimMIM & 0.4181 & 0.6797 & 0.5285 & 0.5421 \\
    ConvMAE & 0.4172 & 0.7276 & 0.5208 & 0.5552 \\
    BootMAE & 0.4133 & 0.7211 & 0.4949 & 0.5431 \\
    SelfMedMAE & 0.4214 & 0.7182 & 0.5173 & 0.5523 \\
    Ours & \textbf{0.4476} & \textbf{0.7351} & \textbf{0.5459} & \textbf{0.5762} \\
    \bottomrule
    \end{tabularx}
    \vspace{-0.5em}
\end{table}

\begin{table}[!t]
    \caption{Performance comparison between AMLP and No New-Net under varying proportions of labeled data and different backbone architectures. }
    \centering
    \begin{tabular}{cc}
        \toprule
        \multicolumn{1}{c}{Methods} & \multicolumn{1}{c}{DSC}   \\
        \midrule
        AMLP (50\%, 2D U-Net,) & 0.7545  \\
        AMLP (100\%, 2D U-Net) & 0.8376  \\
        \midrule
        No New-Net (100\%, 3D U-Net) \cite{isensee2019no} & 0.8210  \\
        AMLP (100\%, 3D U-Net) & 0.8513  \\
        \bottomrule
    \end{tabular}
    \label{tab:no_newnet}
    \vspace{-1.5em}
\end{table}

\noindent\textbf{Analysis of multimodality.}
Theoretically, combining multiple modalities provides complementary visual information to the model, which can help capture the anatomical and pathological features present in medical images more comprehensively and accurately. To this end, besides training on single modalities for BraTS (T1CE, T2, Flair, and T1), we also conducted a new set of experiments where all four modalities were used as input for training simultaneously.

Specifically, we first compared the performance of the proposed AMLP and various self-supervised segmentation methods when all four modalities were used as input for training simultaneously. As shown in Fig. \ref{fig:diff_modal} (a), our proposed model maintained the best performance under 5\%, 10\%, and 50\% annotation ratios when trained with all four modalities, demonstrating the effectiveness of our method. Subsequently, we further investigated the performance difference between training with all four modalities simultaneously and training with each of the four modalities individually. As shown in Fig. \ref{fig:diff_modal} (b), the method of training with all four modalities simultaneously achieved much higher performance than single-modality training across various annotation ratios, indicating that multimodality combination can indeed significantly enhance the model's ability to recognize complex lesions. Therefore, in future work, we will continue to explore strategies for multimodality information fusion to further improve the model's robustness and adaptability.

\begin{table}[!t]
    \caption{Comparison of results for normal organ segmentation on the ABD110 dataset with 10\% labeled data.
    }
    \centering
    \label{normal_organ}
    \setlength{\tabcolsep}{2.2pt} % 
    \begin{tabularx}{0.45\textwidth}{lccccccc}
    \toprule
    Methods & U-Net & MoCo & DenseCL & G-L & MAE & MaskFeat & Ours \\
    \midrule
    DSC &  0.7438 & 0.8135  & 0.8107 & 0.8139 & 0.7767 & 0.7639 & 0.8315 \\
    \bottomrule
    \end{tabularx}
    \vspace{-0.5em}
\end{table}

\begin{figure}[!t]
	\centering
	\subfigure[Various methods trained with multimodal data]{
		\begin{minipage}[b]{0.226\textwidth}
			\includegraphics[width=1\textwidth]{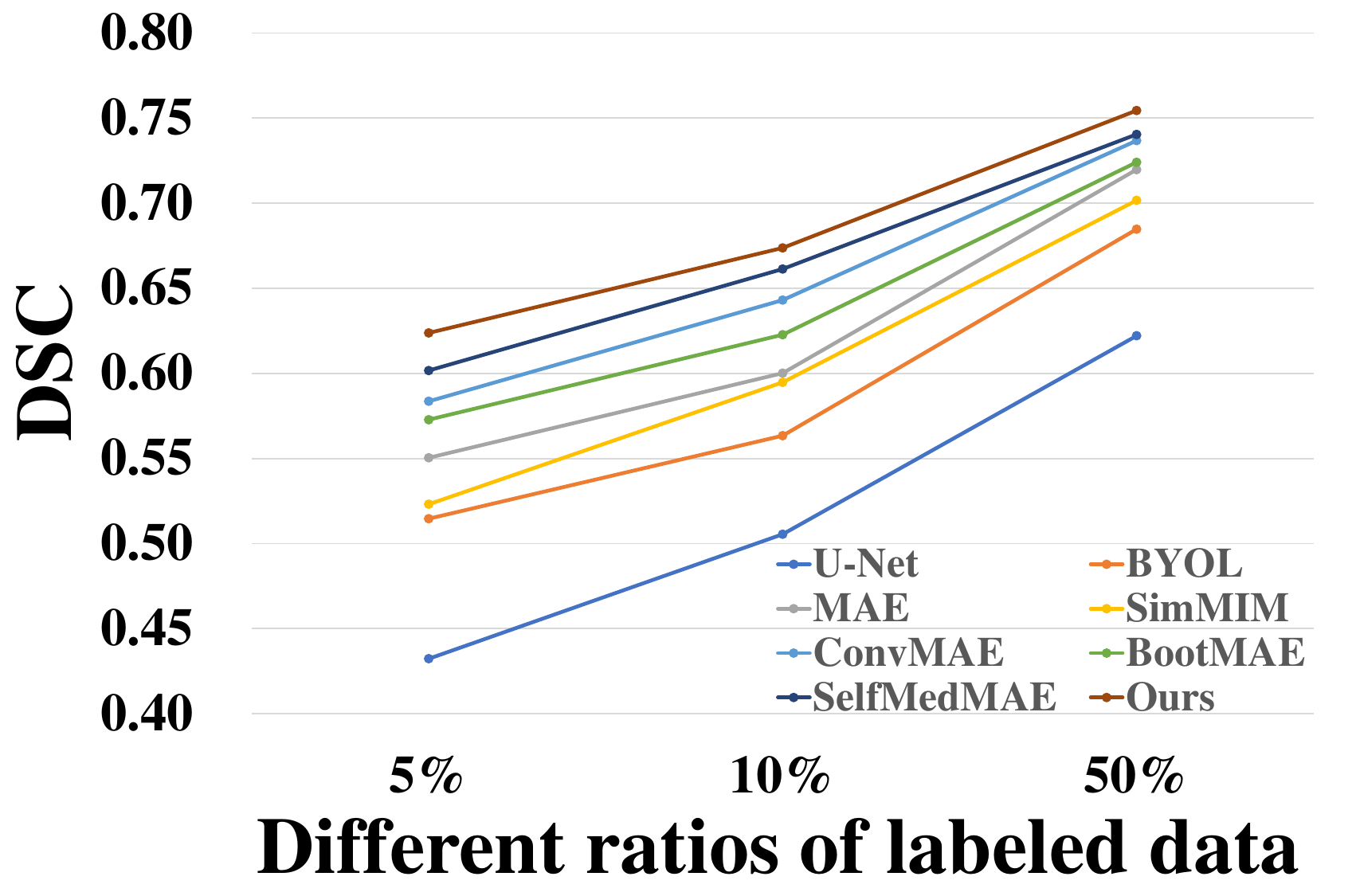}
		\end{minipage}
		% \label{fig:size_hecktor}
	}
    	\subfigure[Our method trained with different modalities data]{
    		\begin{minipage}[b]{0.229\textwidth}
   		 	\includegraphics[width=1\textwidth]{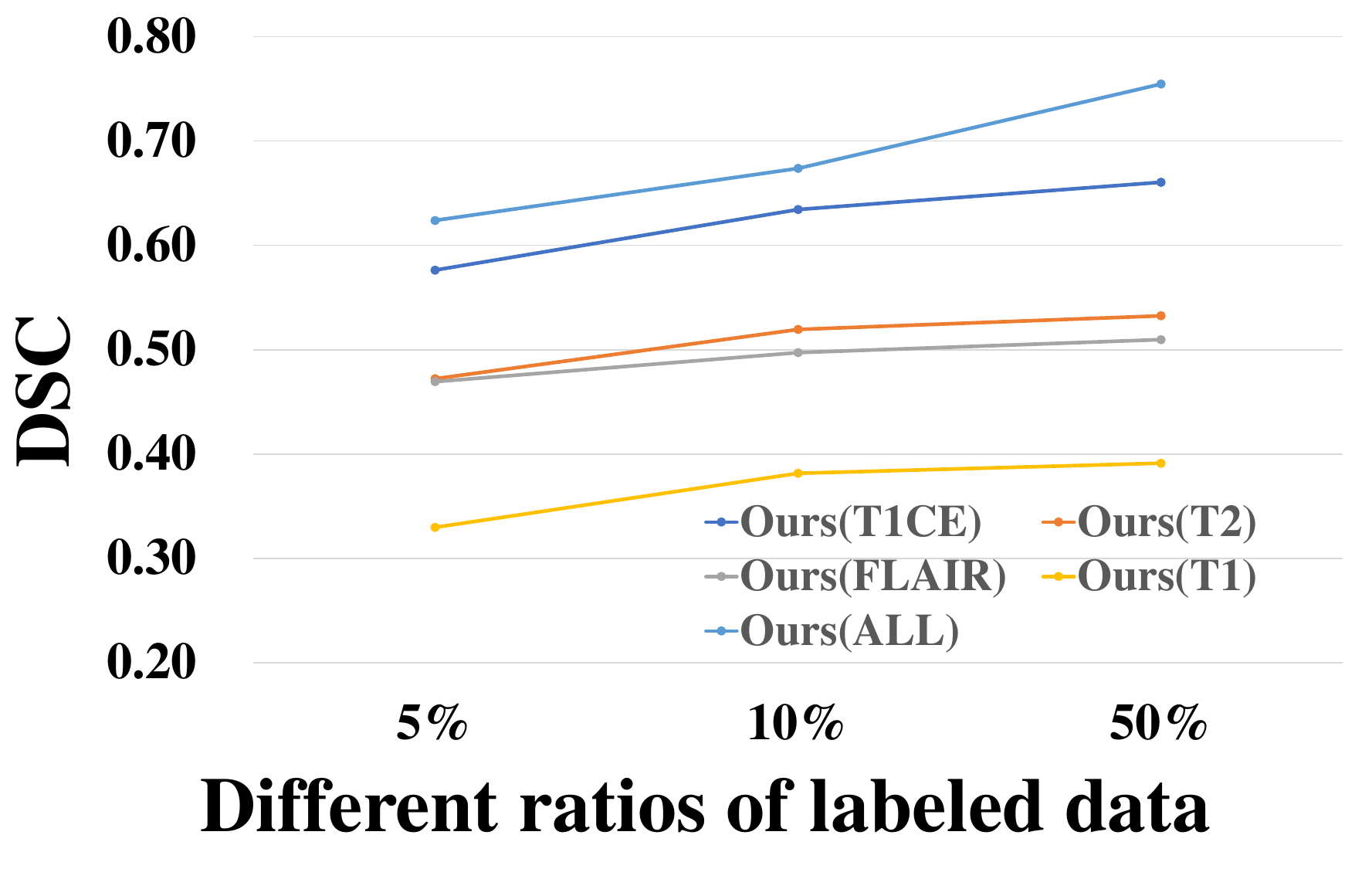}
    		\end{minipage}
		% \label{fig:size_BraTS2018}
    	}
	\caption{Performance comparison of the model using multi-modalities of BraTS2018 with different annotation ratios.}
	\label{fig:diff_modal}
     \vspace{-1.5em}
\end{figure}

\begin{table*}[!t] 
    \centering
    % \footnotesize % 降低字体大小
    \caption{Detection results of the proposed method and baselines on FLAIR modality of BraTS2018 and RSNA Pneumonia dataset with 10\% labeled data.}
    \label{table:detection task}
        % \resizebox{\linewidth}{!}{
        \begin{tabularx}{0.75\textwidth}{XlXXXXXXXXX}
        % \begin{tabularx}{0.85\textwidth}{llllllllll}
            \toprule
            \multicolumn{2}{c}{\multirow{2}*{\textbf{Methods}}}  & \multicolumn{4}{c}{\textbf{BraTS2018}} & \multicolumn{4}{c}{\textbf{RSNA Pneumonia}} \\ \cmidrule(lr){3-6} \cmidrule(lr){7-10} 
            % & & \multicolumn{2}{c}{AP$\uparrow$} & \multicolumn{2}{c}{Sen$\uparrow$} \\ \cmidrule(lr){3-4} \cmidrule(lr){5-6} 
            & & {$AP_{50}$} & {$AP_{70}$}  & {$R_{50}$} & {$R_{70}$} & {$AP_{50}$} & {$AP_{70}$}  & {$R_{50}$} & {$R_{70}$} \\ \midrule
            
            \multirow{5}{*}{\textbf{10\%}}{} &{Faster-RCNN}
            &{0.4301}&{0.1567}
            &{0.4978}&{0.1929}
            &{0.384}&{0.0991}
            &{0.5052}&{0.2129}
            \\

            \multicolumn{1}{c}{}&{BYOL}
            &{0.4523}&{0.2105}
            &{0.5163}&{0.2647}
            &{0.3911}&{0.0997}
            &{0.5062}&{0.2159}
            \\
            
            \multicolumn{1}{c}{}&{MAE}
            &{0.4772}&{0.2361}
            &{0.5451}&{0.2854}
            &{0.3937}&{0.1016}
            &{0.5104}&{0.2260}
            \\

            \multicolumn{1}{c}{}&{ConvMAE}
            &{0.4904}&{0.2526}
            &{0.5507}&{0.3034}
            &{0.3945}&{0.1023}
            &{0.5165}&{0.2321}
            \\

            \multicolumn{1}{c}{} &{Ours} 
            &{\textbf{0.5281}}&{\textbf{0.2812}}
            &{\textbf{0.5763}}&{\textbf{0.3345}}
            &{\textbf{0.4107}}&{\textbf{0.1045}}
            &{\textbf{0.5211}}&{\textbf{0.2354}}
            \\   \midrule

            \multirow{1}{*}{\textbf{50\%}}{} &{Faster-RCNN} 
            &{0.5319}&{0.3057}
            &{0.5909}&{0.3684}
            &{0.4023}&{0.1055}
            &{0.5239}&{0.2287}
            \\   \midrule

            \multirow{1}{*}{\textbf{100\%}}{} &{Faster-RCNN} 
            &{0.6045}&{0.3821}
            &{0.6792}&{0.4593}
            &{0.4789}&{0.1241}
            &{0.6313}&{0.2629}
            \\    \bottomrule
            
        \end{tabularx}
        % }
    \vspace{-1em}
\end{table*}

We also noted that the multimodal segmentation results in Fig. \ref{fig:diff_modal} are not as high as those reported for the BraTS dataset in other papers (e.g., those in \cite{isensee2019no}). This is because our work is based on a self-supervised approach, where only a certain proporition of the data has segmentation labels. In contrast, many multimodal segmentation studies are conducted in a fully-supervised setting, utlizing complete segmentation annotation for all data. Additionally, we employ a different backbone model from some works (e.g., \cite{isensee2019no} uses 3D U-Net, while we use 2D U-Net). Therefore, to demonstrate the superior performance of our proposed AMLP method for multi-modal segmentation and to prevent an underestimation of its capabilities due to setup discrepancies, we use  \cite{isensee2019no} as a baseline and evaluate AMLP in a fully-supervised setting, with both 2D-UNet and 3D-UNet as backbones. The results are presented in Table \ref{tab:no_newnet}. We can observe that under the same 3D U-Net and 100\% labeled data, our model achieves 0.8513, surpassing No New-Net by 0.303. This result robustly demonstrates AMLP imparts a substantial and inherent advantage, enabling the learning of more effective representations.

\begin{table}[!t]
    \caption{Hyperparameter configuration for model training.}
    \centering
    \label{table:hyperparameters}
    \begin{tabularx}{0.4\textwidth}{Xl}
    \toprule
    config & value  \\
    \midrule
    optimizer & [SGD, Adam, \underline{AdamW}] \\
    base learning rate & [1e-5, \underline{1e-4}, 1e-3] \\
    weight decay & [0.005, 0.01, \underline{0.05}, 0.1] \\
    optimizer momentum & $\beta_1$; $\beta_2$=0:9; 0:95 \\
    batch size & 512 \\
    learning rate schedule & cosine decay \\
    warmup epochs & [20, \underline{40}, 60] \\
    training epochs & [400, \underline{800}, 1200, 1600] \\
    augmentation & RandomResizedCrop \\
    \bottomrule
    \end{tabularx}
    \vspace{-1em}
\end{table}

\begin{table*}[t]
  \centering
  \caption{Comparison of 2D and 3D segmentation methods on the BraTS2018 multimodal dataset.}
  \label{tab:comparison}
  \renewcommand{\arraystretch}{1.1}
    \begin{tabularx}{0.67\textwidth}{c c >{\centering\arraybackslash}X c c}
    \toprule
    \multirow{2}{*}{} & \multirow{2}{*}{Backbone} & \multirow{2}{*}{Method} & \multicolumn{2}{c}{Metrics} \\ \cline{4-5}
    & & & DSC $\uparrow$ & HD95 $\downarrow$ \\ \midrule
    \multirow{8}{*}{2D}
       & \multirow{2}{*}{CNN}
       & MAE (U-Net), w/o ours & 0.8182 & 7.47\\
       & & MAE (U-Net), w/ ours  & 0.8376 & 6.26\\ \cline{2-5}
       & \multirow{4}{*}{Transformer}
       & MAE (ViT), w/o ours      & 0.8103 & 9.24\\
       & & MAE (ViT), w/ ours      & 0.8309 & 7.32\\
       & & SimMIM (Swin), w/o ours & 0.8250 & 7.27\\
       & & SimMIM (Swin), w/ ours  & 0.8414 & 5.62\\ \cline{2-5}
       & \multirow{2}{*}{Mamba}
       & U-Mamba, w/o ours        & 0.8495 & 6.93\\
       & & U-Mamba, w/ ours        & 0.8662 & 5.72\\ \midrule
    \multirow{8}{*}{3D}
       & \multirow{2}{*}{CNN}
       & MAE (3D U-Net), w/o ours & 0.8337 & 7.29\\
       & & MAE (3D U-Net), w/ ours & 0.8513 & 5.63\\ \cline{2-5}
       & \multirow{4}{*}{Transformer}
       & SelfMedMAE (ViT), w/o ours & 0.8294 & 8.06\\
       & & SelfMedMAE (ViT), w/ ours & 0.8481 & 6.22\\
       & & ConvMAE (CNN+Trans.), w/o ours & 0.8325 & 7.84\\
       & & ConvMAE (CNN+Trans.), w/ ours  & 0.8502 & 6.71\\ \cline{2-5}
       & \multirow{2}{*}{Mamba}
       & SegMamba, w/o ours        & 0.8505 & 5.63\\
       & & SegMamba, w/ ours        & 0.8756 & 4.74\\
    \bottomrule
  \end{tabularx}
\end{table*}

\noindent\textbf{Analysis of normal anatomical structures segmentation tasks.}
Although the components of AMLP were originally designed for the lesion segmentation scenario, their working mechanism gives them the potential to handle a wider range of medical image analysis tasks. For example, MPS operates based on the principle of binary classification in feature space, which is not only suitable for segmenting lesion and non-lesion areas, but also for segmenting organs and background. RRL and CCL focus on extracting and optimizing unique feature representations of target categories, regardless of whether they are lesions or organs.

To further explore the applicability of AMLP to the normal anatomy segmentation task, we performed abdominal segmentation experiments on the publicly available ABD110 dataset with 10\% labeled data.
The results are shown in Table. \ref{normal_organ}. 
We can observe that our proposed method achieves the best performance among all methods, with a DSC 1.76\% higher than the second-best result, which benefits from the model's learning of key discriminative features for distinguishing anatomical regions from backgrounds. These results indicate that our proposed method can be used not only for lesion segmentation tasks, but also for normal anatomy segmentation tasks.

\noindent\textbf{Hyperparameters for pre-training.}
To ensure the robustness and reproducibility of our results, we performed a systematic hyperparameter optimization for the pre-training stage. We employed a step-wise grid search methodology, and the search space for key hyperparameters is detailed in Table \ref{table:hyperparameters}, with the final optimal configuration underlined. Our search results reveal that the AdamW optimizer, combined with a learning rate of 1e-4 and a weight decay of 0.05, yielded the most effective feature representations for downstream tasks. All results are the average of three independent runs using this optimal setup.
Furthermore, we have evaluated the performance of training duration, as shown in Fig. \ref{fig:epoch}. The downstream performance (DSC) on the BraTS2018 T1 modality (10\% labels) plateaus around 800 epochs, indicating that longer training provides diminishing returns and risks overfitting to the pre-training task. We thus selected 800 epochs as the optimal trade-off between performance and efficiency.

\noindent\textbf{Analysis of detection task.}
In order to more comprehensively demonstrate AMLP's ability to learn and represent lesion features, as well as its wide impact and applicability to downstream tasks, we have conducted additional medical image lesion detection experiments.

Specifically, we verified the performance of AMLP on detection task with the public BraTS ~\cite{Brats2018_1} and RSNA Pneumonia ~\cite{shih2019augmenting} detection dataset. BraTS detection dataset is converted from the FLAIR modality of BraTS segmentation dataset, where segmentation labels are converted to detection labels. RSNA Pneumonia detection dataset is a public medical imaging dataset for pneumonia detection created by the Radiological Society of North America (RSNA).
In order to better evaluate the performance of AMLP in detection tasks, we used the most widely used average precision (AP) and Recall (R) as evaluation indicators. The detection results of AMLP at different annotation ratios are compared with contrastive learning method (BYOL \cite{grill2020bootstrap}), the masked image modeling method (MAE \cite{he2022masked} , ConvMAE \cite{gao2022convmae}), and Faster-RCNN \cite{ren2015faster} as shown in Table \ref{table:detection task}.

\begin{figure}[!t]
	\centering
    \begin{minipage}{0.45\textwidth}
    \includegraphics[width=\textwidth]{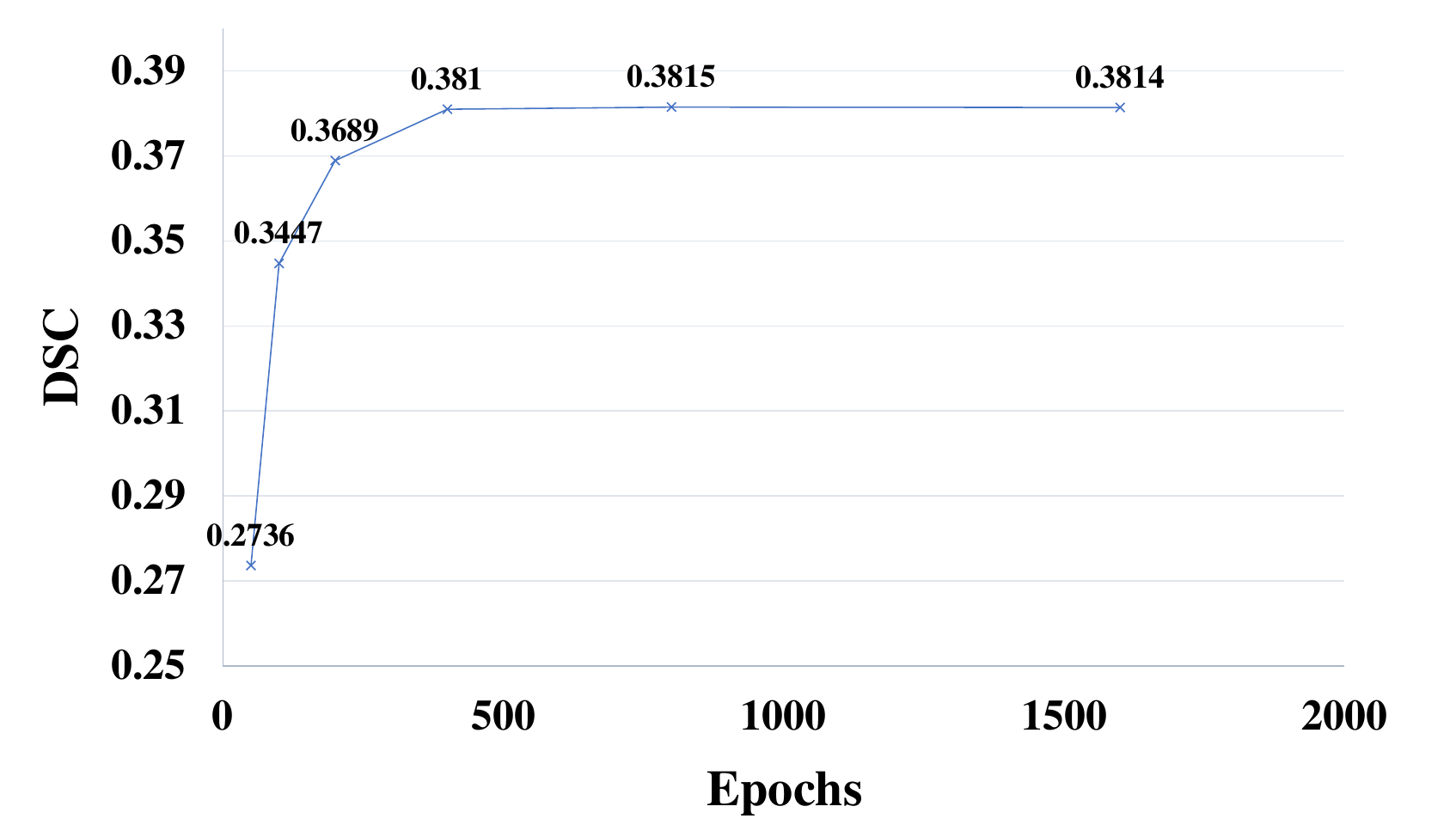}
    \vspace{-1.5em}
    \end{minipage}
	\caption{Training schedules. The DSC results of training our proposed AMLP trained at different epochs on the T1 modality of BraTS2018 dataset with 10\% labeled data.}
	\label{fig:epoch}
    \vspace{-1.5em}
\end{figure}

\begin{table*}[!t]
    \caption{Comparison of efficiency between vision foundation models and AMLP on the BraTS2018 multimodal dataset.}
    \centering
    \label{resource_comparison}
    % \resizebox{\columnwidth}{!}{%
    \begin{tabular}{lcccccc}
    \toprule
    Methods & AMLP  & SAM & SAM v2 & Med-SAM & MA-SAM & 3D SAM-Adapter \\
    \midrule
    Total Parameters (M)  & 31  & 636 & 224 &  95 & 97 & 102.4 \\
    \midrule
    Inference Peak Memory (GB) & 0.4  & 5.2 & 3.5 &  2.3  & 2.4 & 8.7 \\
    \bottomrule
    \end{tabular}
    \vspace{-1em}
\end{table*}

\begin{table}[!h]
    \caption{\color{DeepPink}{Comparison of segmentation results of vision foundation models and AMLP on BraTS2018 multimodal dataset.}}
    \centering
    \label{2D_foundation_models}
    \begin{tabular}{llcc}
    \toprule
    Methods  & Initialization weights & DSC$\uparrow$ & BIoU$\uparrow$ \\
    \midrule
    AMLP (50\%)        & --             & 0.7545 & 0.3713 \\
    AMLP (100\%)       & --             & 0.8376 & 0.4782 \\
    \midrule
    SAM                & ViT-H          & 0.7422 & 0.3881 \\
    SAM v2             & Hiera-L        & 0.7610 & 0.4079 \\
    \midrule
    MedSAM             & MedSAM-ViT-B   & 0.8067 & 0.4505 \\
    MA-SAM             & SAM-ViT-B          & 0.8010 & 0.4474 \\
    \color{DeepPink}{MA-SAM}           & \color{DeepPink}{MedSAM-ViT-B}   & \color{DeepPink}{0.8125} & \color{DeepPink}{0.4570} \\
    3D SAM-Adapter     & SAM-ViT-B          & 0.8153 & 0.4592 \\
    \color{DeepPink}{3D SAM-Adapter}   & \color{DeepPink}{MedSAM-ViT-B}   & \color{DeepPink}{0.8271} & \color{DeepPink}{0.4682} \\
    \bottomrule
    \end{tabular}
    \vspace{-1em}
\end{table}

As observed, various self-supervised methods outperform Faster R-CNN in detection accuracy when using only 10\% labeled data. Additionally, our proposed AMLP method achieved the best performance among all self-supervised methods, nearing the performance of Faster R-CNN detection model at 50\% annotation.
On BraTS dataset, compared to the second-best baseline method, AMLP achieved 3.77\% better results on $AP_{50}$ and 2.56\% better results on $R_{50}$. On RSNA Pneumonia detection dataset, AMLP outperformed the second-best baseline by 1.62\% on $AP_{50}$ and 0.46\% on $R_{50}$. These findings strongly indicate that our method has successfully generalized in the task of lesion detection. AMLP learns more discriminative features from masked lesions, which significantly benefits the detection task.

\noindent\textbf{Extensibility to Different Backbones.}
To validate the applicability of our strategy on different architectures and dimensionality settings, besides using U-Net as the backbone, we have also conducted comprehensive experiments across different backbones including 2D and 3D and various architectures including CNN-based (U-Net \cite{ronneberger2015u}, 3D U-Net \cite{cciccek20163d}), transformer-based (MAE \cite{he2022masked}, SimMIM \cite{xie2022simmim}, SelfMedMAE \cite{goncharov2023vox2vec}, ConvMAE \cite{gao2022convmae}), and the recent Mamba-based models (U-Mamba \cite{ma2024u}, SegMamba \cite{xing2024segmamba}). The results are shown in Table \ref{tab:comparison}. We can observe that our proposed method (\textit{w/ ours}) consistently outperforms the baseline (\textit{w/o ours}) across all evaluated architectures. Specifically, we observe a notable increase in DSC (nearly 2\% on average) and a reduction in HD95 (about 1.5 on average). These consistent improvements across CNN-based, transformer-based, and Mamba-based architectures, in both 2D and 3D formulations, validate the effectiveness and robustness of our SSL strategy.

\noindent\textbf{Comparing with Vision Foundation Models.}
As the vision foundation models emerge, we also want to know the performances of our AMLP model comparing to the existing vision foundation models, e.g, SAM \cite{kirillov2023segment}, SAM v2 \cite{ravi2024sam}, Med-SAM \cite{ma2024segment}, MA-SAM \cite{chen2024ma} and 3D SAM-Adapter \cite{gong20243dsam}. However, Firstly, the large model is pre-trained with a large number of images of various types (e.g., 11 million images for SAM), while our model is training from scratch with only hundreds of cases in total, so they have a huge difference in terms of training data; Secondly, the vision foundation models have a large number of parameters, and their training is very computationally and memory-intensive and time-consuming; Therefore, it is very unfair to directly compare the segmentation results obtained by our self-supervised training based on only 10\% of the labeled data with the results of the medical segmentation foundation models trained with massive data. Therefore, we adjust our model training settings and train AMLP using multimodal BRATS data with 50\% and 100\% annotations respectively, so as to try to bridge the data gap between AMLP and the foundation models, and then compare the performance differences between AMLP and the large model under these settings.

\color{DeepPink}{As shown in Table \ref{2D_foundation_models},} our AMLP outperforms all five vision foundation models with 100\% labeled data. For example, with 100\% labels, AMLP outperforms MA-SAM (based on SAM-ViT-B initialization weights) by 3.66\% in DSC, and 3.08\% in BIoU. Besides, we also compare with MA-SAM \cite{chen2024ma} and 3D SAM-Adapter \cite{gong20243dsam} initialized with MedSAM-ViT-B weights \cite{ma2024segment}: AMLP outperforms MA-SAM with MedSAM-ViT-B by 2.51\% in DSC and 2.12\% in BIoU; while outperforming 3D SAM-Adapter by 1.05\% in DSC and 1.00\% in BIoU.
Even with 50\% of labeled data, AMLP remains competitive with baselines like Med-SAM \cite{ma2024segment}. This indicates that, with sufficient domain-specific data, our specific pre-training strategy attains segmentation accuracy comparable to or better than directly using or parameter-efficiently adapting generalist foundation models.
Furthermore, as shown in Table \ref{resource_comparison}, our proposed AMLP has only 31 MB total parameters and 0.4 GB inference peak memory. The parameters of foundation models are 3 or even 20 times that of AMLP, and inference peak memory is 6 or even 22 times that of AMLP. Besides, replacing the SAM-ViT-B weights with MedSAM-ViT-B for MA-SAM and 3D SAM-Adapter does not change the number of parameters and inference peak memory, because they use the same SAM ViT-B backbone. In conclusion, our AMLP model can achieve better segmentation results with limited computational resources.}

\section{Discussion and future work}
\subsection{Social impact for proposed algorithm}
The proposed model has wide-ranging applications in daily clinical scenarios, significantly reducing the burden on specialists while safeguarding diagnostic precision. For example, in cancer radiotherapy, precisely delineating tumors in 3D CT or PET scans often requires hours of slice-by-slice annotation and the collaboration of multiple physicians—a task that consumes limited human resources and extends patient waiting times for treatment. In this context, our method can generate a high-quality initial contour in seconds, allowing clinicians to then refine the lesion, thereby streamlining the clinical annotation workflow. Similarly, in multi-parametric brain magnetic resonance imaging (MRI), our network can generate high-fidelity glioma boundaries to assist neurosurgeons in designing minimally invasive surgical trajectories, thus accelerating the entire treatment planning process.

Furthermore, our self-supervised framework achieves precise segmentation without the high annotation costs of fully supervised methods. This greatly reduces the data requirements for automated medical image segmentation systems in clinical settings, helps improve their deployment efficiency, and ultimately lowers patient examination costs in specific clinical scenarios.

\vspace{-0.5em}

\subsection{Limitations and future work}
Although our proposed AMLP solves the problem of MIM application deviation in natural images and medical images to a certain extent. It still have some improvement space. For instance, we can significantly improve the mask selection strategy by integrating domain-specific expertise and prior knowledge from the field of medicine. This could involve considering prevalent lesion locations associated with various diseases and accounting for the distinct morphological characteristics exhibited by different types of lesions. Employing such tailored mask selection techniques holds substantial promise in enhancing the model's ability to discern between various diseases and lesions, thus contributing to the overall improvement of medical image segmentation.

While our method demonstrates superior performance in comparison to existing self-supervised methods, there remains a little performance gap when comparing to fully supervised methods. It can potentially be addressed through incorporation of more advanced techniques like reinforcement learning \cite{Yunxin-TMI2024, TNNLS2023}. Specifically, the mask selection process can be conceptualized as a Markov decision process, enabling adaptive optimization through policy gradient methods. This approach may provide a more effective solution.

Moreover, although we have trained our model on various datasets encompassing different modalities (such as CT and MRI) and anatomical regions (such as head and neck cancer, and brain tumors), its performance may not maintain the same level of stability and accuracy when transferred to heterogeneous datasets with different sources (e.g., from different institutions or acquired using different equipment) or with inconsistent feature distributions. This limitation may lead to suboptimal performance when the model encounters unseen data, thereby affecting its applicability and robustness in broader clinical environments. Therefore, future research should incorporate more diverse datasets and employ techniques such as transfer learning to enhance the model's generalization capabilities and improve its adaptability across different application scenarios.

In summary, combining cross-knowledge with deep learning methods stands as a key direction for future medical image segmentation research. Such integrated approaches have potential to enhance model performance, and provide more reliable support for patient health.

\vspace{-0.5em}
\section{Conclusion}

In this paper, we proposed a self-supervised framework named Adjustable Masked Lesion Patches (AMLP) for medical image segmentation. This architecture can leverage unlabeled medical images for representation learning by masking and reconstructing lesion patches. Specifically, under the guidance of the masked patch selection strategy, the model can focus on learning informative representations of important lesion areas. In addition, we designed an relative reconstruction loss and category consistency loss, which can strengthen the model's focus on lesions patches and enhance distinction between lesions and background. Moreover, we developed an adjustable masking ratio strategy that gradually increases the masking ratio over training to expand the scope of learnable mutual information. Through extensive experiments on two public datasets, we illustrate that our method can achieve remarkable performance with only a small portion of annotations. This capability substantially enhances the utilization of the wealth of unlabeled medical images.

% \vspace{-0.5em}

% \section*{Acknowledgment}
% The preferred spelling of the word ``acknowledgment'' in American English is 
% without an ``e'' after the ``g.'' Use the singular heading even if you have 
% many acknowledgments. Avoid expressions such as ``One of us (S.B.A.) would 
% like to thank $\ldots$ .'' Instead, write ``F. A. Author thanks $\ldots$ .'' In most 
% cases, sponsor and financial support acknowledgments are placed in the 
% unnumbered footnote on the first page, not here.

% \bibColoredItems{DarkRed}{zhou2021preservational}
% \bibColoredItems{DarkRed}{shih2019augmenting}
% \bibColoredItems{DarkRed}{ren2015faster}
% \bibColoredItems{MediumBlue}{zhou2023self}
% \bibColoredItems{DarkOrange}{isensee2019no}
% \bibColoredItems{DarkOrange}{Li2024SelfSupervisedAL}
% \bibColoredItems{DarkOrange}{seince2024dense}
% \bibColoredItems{DeepPink}{ma2024u}
% \bibColoredItems{DeepPink}{gong20243dsam}
% \bibColoredItems{DeepPink}{chen2024ma}
% \bibColoredItems{Pink}{}
% \bibColoredItems{Green}{dice1945measures}
% \bibColoredItems{MediumBlue}{DHANACHANDRA2015764}
% \bibColoredItems{Green}{jerri1977shannon}
% \bibColoredItems{Green}{wang2008probability}
% \bibColoredItems{Green}{shannon1948mathematical}
% \bibColoredItems{Green}{neal2012bayesian}
% \bibColoredItems{Green}{robert1999monte}
% \bibColoredItems{Green}{cover1999elements}
% \bibColoredItems{Green}{bishop2006pattern}
% \bibColoredItems{Green}{verma2022uniform}
% \bibColoredItems{Green}{wettig2022should}
% \bibColoredItems{Green}{Yunxin-TMI2024}
\bibliography{refs.bib}
\bigskip

\end{document}